\documentclass[10pt,twocolumn,letterpaper]{article}

\usepackage[pagenumbers]{cvpr} % To force page numbers, e.g. for an arXiv version

\usepackage{xspace}
\usepackage{colortbl} % 用于添加背景颜色
\usepackage{xcolor}   % 定义颜色

\def\modelname{SynerGen-VL\xspace}

\definecolor{cvprblue}{rgb}{0.21,0.49,0.74}
\definecolor{prompt}{rgb}{0.21,0.49,0.74}
\usepackage[pagebackref,breaklinks,colorlinks,allcolors=cvprblue]{hyperref}
\usepackage{multirow}

\usepackage{hyperref}
\usepackage{multirow}
\usepackage{url}
\usepackage{tabularx}
\usepackage{array} 
\usepackage{enumitem}
\usepackage{threeparttable}
\usepackage{graphicx} % for pdf, bitmapped graphics files
\usepackage{amsmath} % assumes amsmath package installed
\usepackage{amssymb}  % assumes amsmath package installed
\usepackage{algorithmic}
\usepackage{algorithm}
\usepackage{caption}
\usepackage{wrapfig}
\usepackage{booktabs}
\usepackage{colortbl}
\usepackage{pifont}
\usepackage{xcolor}
\usepackage{float}

\usepackage{marvosym}
\usepackage{mdframed}
\usepackage{tcolorbox}
\usepackage{xcolor}
\definecolor{mylightblue}{RGB}{100,149,237} % 定义一个自定义的浅蓝色，这里使用的是LightBlue的RGB代码

\definecolor{darkgreen}{rgb}{0.0, 0.5, 0.0} % 您可以调整 RGB 值来控制颜色的深浅
\definecolor{mambaPurple}{rgb}{197, 153, 204}
\usepackage{CJK}
\usepackage{enumitem}
\usepackage{pifont}

\def\paperID{*****} % *** Enter the Paper ID here
\def\confName{****}
\def\confYear{2025}

\title{SynerGen-VL: Towards Synergistic Image Understanding and Generation with Vision Experts and Token Folding}

\author{
    Hao Li$^{1,2*\dag}$,
    Changyao Tian$^{2,1*\dag}$,
    Jie Shao$^{3,1*\dag}$,
    Xizhou Zhu$^{4,5*}$,
    Zhaokai Wang$^{6,1\dag}$,
    Jinguo Zhu$^{1}$, \\
    Wenhan Dou$^{4,5}$, 
    Xiaogang Wang$^{2}$, 
    Hongsheng Li$^{2}$,
    Lewei Lu$^{5}$,
    Jifeng Dai$^{4,1,7}$\textsuperscript{\Letter}\vspace{0.5em}  \\ 
    \fontsize{11pt}{12pt}\selectfont
    $^1$OpenGVLab, Shanghai AI Laboratory~~~
    $^2$MMLab, The Chinese University of Hong Kong~~~ \\
    \fontsize{11pt}{12pt}\selectfont
    $^3$Nanjing University~~~
    $^4$Tsinghua University~~~
    $^5$SenseTime Research~~~
    $^6$Shanghai Jiao Tong University\\
    \fontsize{11pt}{12pt}\selectfont
    $^7$Beijing National Research Center for Information Science and Technology
}

\newcommand\blfootnote[1]{%
\begingroup
\renewcommand\thefootnote{}\footnote{#1}%
\addtocounter{footnote}{-1}%
\endgroup
}

\begin{document}
\maketitle
\begin{abstract}
    The remarkable success of Large Language Models (LLMs) has extended to the multimodal domain, achieving outstanding performance in image understanding and generation. Recent efforts to develop unified Multimodal Large Language Models (MLLMs) that integrate these capabilities have shown promising results. However, existing approaches often involve complex designs in model architecture or training pipeline, increasing the difficulty of model training and scaling.
    In this paper, we propose \modelname, a simple yet powerful encoder-free MLLM capable of both image understanding and generation. To address challenges identified in existing encoder-free unified MLLMs, we introduce the token folding mechanism and the vision-expert-based progressive alignment pretraining strategy, which effectively support high-resolution image understanding while reducing training complexity.
    After being trained on large-scale mixed image-text data with a unified next-token prediction objective, \modelname achieves or surpasses the performance of existing encoder-free unified MLLMs with comparable or smaller parameter sizes, and narrows the gap with task-specific state-of-the-art models, highlighting a promising path toward future unified MLLMs. Our code and models shall be released.

    \blfootnote{* Equal contribution. \dag~Interns at Shanghai AI Laboratory.} \blfootnote{\Letter~Corresponding to Jifeng Dai \textless daijifeng@tsinghua.edu.cn\textgreater.}
\end{abstract}
    
\section{Introduction}
\label{sec:intro}

\begin{figure*}[t]
  \centering
  \includegraphics[width=\linewidth]{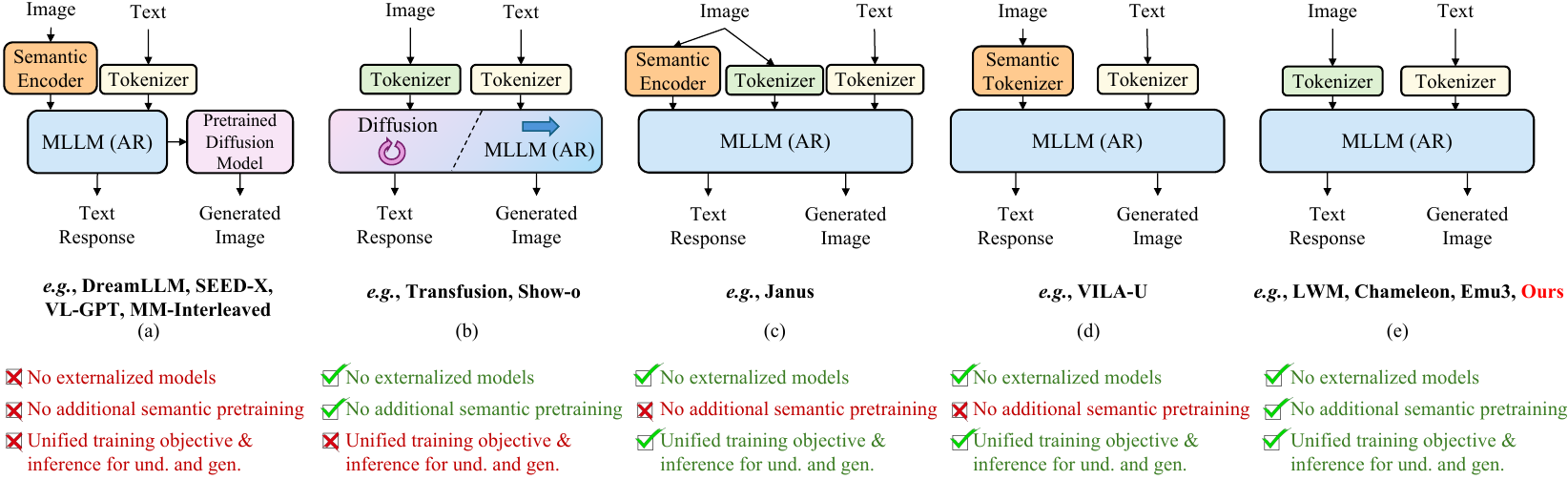}
   \caption{\textbf{Comparison among exemplary unified MLLMs for synergizing image understanding and generation tasks.} Compared with methods (a)$\sim$(d) that incorporate complicated designs of model architectures, training methods, and the use of external pretrained diffusion models, (e) encoder-free unified MLLMs adopt a simple design that uses the simple next token prediction framework for both images understanding and generation tasks, allowing for broader data distribution and better scalability.} 
   \label{fig:intro1}
\end{figure*}

The remarkable success of Large Language Models (LLMs)~\cite{VLM:GPT-4, touvron2023llama, cai2024internlm2} has been extended to the multimodal domain, achieving impressive performance in image understanding~\cite{VLM:GPT-4v, VLM:Gemini, VLM:InternVL-1.5, VLM:LLaVA} and image generation~\cite{llamagen, xiao2024omnigen, var}. 
Recent research has aimed to develop unified Multimodal Large Language Models (MLLMs) with synergistic image understanding and generation capabilities~\cite{dong2023dreamllm, VLM:MM-interleaved, zhou2024transfusion, wu2024janus, wu2024vila, team2024chameleon}. 
Although they have demonstrated competitive performance in both tasks, they often involve complex designs as illustrated in Fig.~\ref{fig:intro1}(a)$\sim$(d),  such as (a) relying on external diffusion models for image generation~\cite{dong2023dreamllm, seed_x, vlgpt, VLM:MM-interleaved}, (b) using different training objectives (\ie diffusion and autoregression) for the two tasks~\cite{zhou2024transfusion, xie2024show}, (c) employing distinct image encoders for each task~\cite{wu2024janus}, and (d) require additional semantic pretraining for image tokenizers~\cite{wu2024vila}. These complexities disrupt the simplicity of the next token prediction paradigm of LLMs, increasing systematic difficulty and limiting scalability.

To address these complexities, some studies have tried to develop unified MLLMs with simple architectures, eliminating dependencies on external models, distinct task-specific models, and additional semantic pretraining~\cite{team2024chameleon, emu3, lwm}. As shown in Fig.~\ref{fig:intro1}(e), these approaches adopt a similar tokenization strategy for both images and text, and model both image understanding and generation tasks within a unified next token prediction framework. 
The image tokenizers~\cite{vqgan} are pretrained for reconstruction on pure image data, without requiring human annotations or text supervision, which allows for a broad data distribution and strong scalability.
These concise and scalable designs have demonstrated a promising path toward synergistic image understanding and generation.

Nevertheless, these methods still face some key challenges in practical use. Specifically, (1) since both image understanding and generation rely entirely on MLLMs, substantial training is required to incorporate vision capabilities into MLLMs. However, this may interfere with the pretrained knowledge of LLMs, resulting in reduced general perception and generalization capabilities. Although existing methods try to avoid this by training MLLM from scratch using mixed text and multimodal data, they face considerable challenges in optimizing stability, data quality, and training cost~\cite{team2024chameleon, emu3}; (2) current visual tokenizers require low feature downsample ratios to ensure reconstruction with fine details~\cite{vqgan}. This results in long visual token sequences for high-resolution images, which is unsuitable to LLMs and limits the use of high-resolution images, thus affecting performance, especially for image understanding.

In this paper, we aim to build a simple yet powerful unified MLLM that addresses the aforementioned challenges.
Specifically, 1) inspired by image understanding models with Multimodal Mixture-of-Experts (MMoE)~\cite{wang2023cogvlm, unimoe, mono_internvl} structure, we introduce vision experts with additional parameters dedicated to image representation. Aligning the vision experts to the frozen LLM helps integrate vision capabilities while minimizing disruption to the LLM's pretrained knowledge; 2) to effectively support high-resolution images, the input visual token sequence can be compressed to reduce its length, while an additional decoder would be employed during image generation to reconstruct detailed image sequences from the compressed representations.

\begin{figure}[ht]
  \centering
  \includegraphics[width=\linewidth]{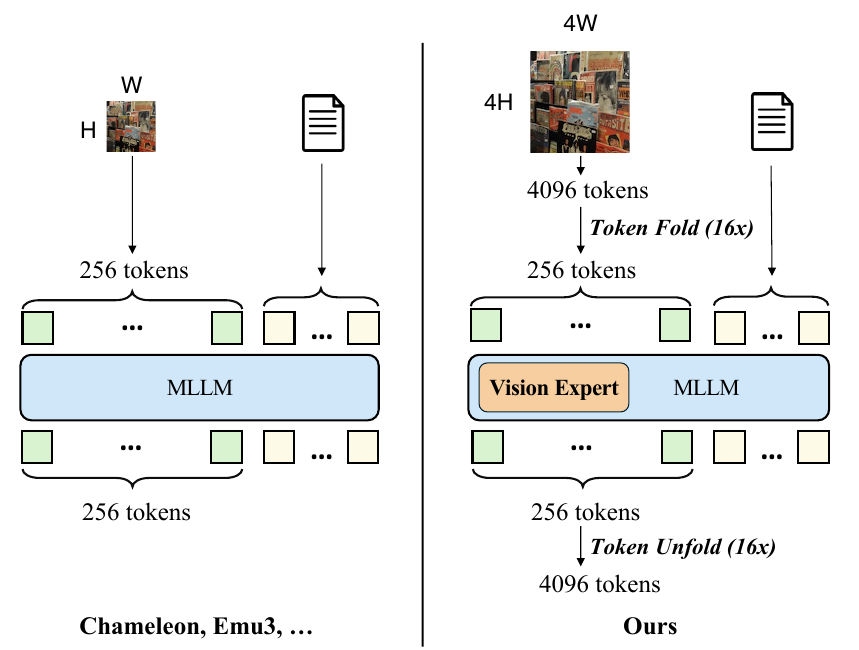}
   \caption{\textbf{Comparision between \modelname and previous encoder-free unified MLLMs.} \modelname adopts a token folding and unfolding mechanism and vision experts to build a strong and simple unified MLLM. With the same image context length, \modelname can support images of much higher resolutions, ensuring the performance of both high-resolution image understanding and generation.
   } 
   \label{fig:intro2}
\end{figure}

Following this perspective, we propose \modelname, a high-performance unified MLLM with synergistic image understanding and generation capabilities, using non-semantic discrete image tokens to represent images. As shown in Fig.~\ref{fig:intro2}, compared with previous encoder-free unified MLLMs, \modelname employs additional vision experts, \ie image-specific Feed-Forward Networks (FFNs), to incorporate vision capabilities into pretrained LLMs. Meanwhile, \modelname uses a hierarchical architecture to increase the feature downsampling ratio within the MLLM. Specifically, the input image token sequences are downsampled by token folding to reduce their lengths. To generate high-quality images, the generated token sequences are unfolded by a shallow autoregressive Transformer head. To preserve the LLM's pretrained knowledge, we perform two-stage alignment pretraining with mixed image understanding and generation data: (1) only image-specific FFNs are trained with noisy web data to achieve basic semantic understanding and image generation aligned with the representation space of LLM; (2) image-specific FFNs and self-attention layers are trained with high-quality image understanding and generation data to further integrate multimodal features into the pretrained LLM. After alignment pretraining, \modelname supports image understanding and generation tasks simultaneously through supervised instruction fine-tuning.

We train \modelname on large-scale mixed image-text data and evaluate it on a range of image understanding and generation benchmarks. Experimental results demonstrate that, with its simple design, \modelname achieves or surpasses the performance of existing encoder-free unified MLLMs with comparable or smaller parameter sizes, and narrows the gap with task-specific state-of-the-art (SoTA) models. In particular, with only 2.4B activated parameters, \modelname achieves image understanding and generation performance on par with Emu3~\cite{emu3}, which has 8B parameters,  highlighting its strong potential as a promising path towards next-generation unified MLLM. Our contributions are summarized as follows:
\begin{itemize}
    \item We propose \modelname, a Multimodal Large Language Model (MLLM) with simple architecture and training process, capable of handling both image understanding and generation through a unified next token prediction paradigm.
    
    \item We introduce the token folding mechanism and the vision-expert-based progressive alignment pretraining to unified MLLMs, which effectively support high-resolution image understanding and reduce training difficulty.

    \item Experiments demonstrate that \modelname achieves competitive performance in a range of image understanding and generation benchmarks, revealing a promising path towards future unified MLLM.
\end{itemize}

\section{Related Work}
\label{sec:related_work}

\begin{figure*}[t]
  \centering
   \includegraphics[width=0.95\linewidth]{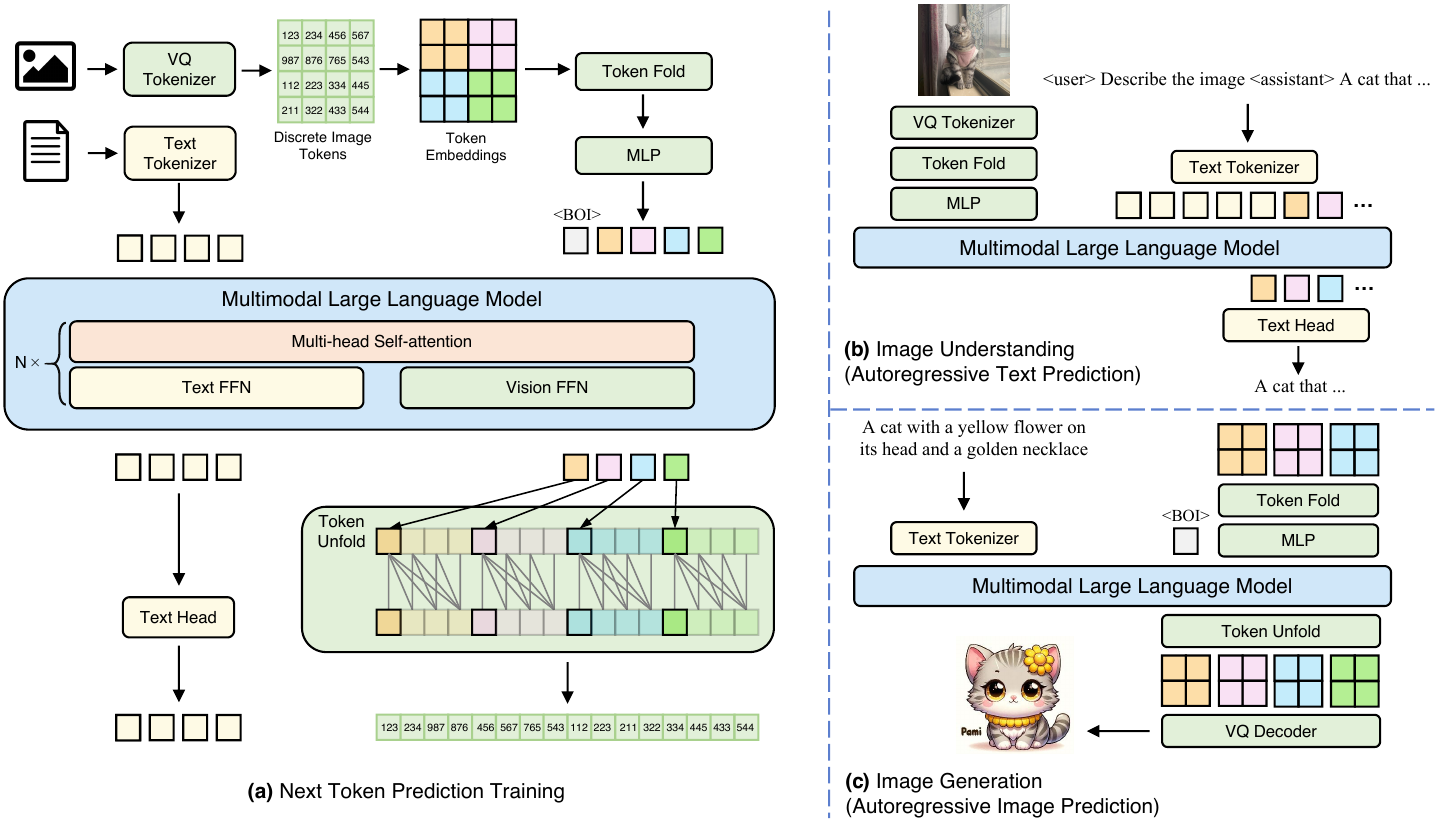}

   \caption{\textbf{Overview of the proposed \modelname.} The image and text are represented as discrete tokens, and modeled with a single LLM and unified next-token prediction paradigm. Text and vision expert FFNs are introduced to incorporate visual capabilities into the pretrained LLM. To support processing high-resolution images, the input image token sequence is folded to reduce its length, and unfolded by a shallow autoregressive Transformer head to generate images.} 
   \label{fig:method}
\end{figure*}

\paragraph{Unified MLLMs for Synergistic Image Understanding and Generation.}

Unifying image understanding and generation in a single MLLM has attracted wide academic attention. Early efforts primarily integrate an external diffusion decoder for image generation~\cite{VLM:EMU, VLM:EMUv2, seed_x, VLM:NeXTGPT, VLM:MiniGemini}.
Inspired by the success of next-token prediction in LLMs, some studies explore using discrete visual tokens to represent and generate images in a fully autoregressive paradigm~\cite{lwm, lavit,  team2024chameleon, cm3leon, wu2024vila, emu3}.
To achieve high performance for both image understanding and generation, some recent methods have decoupled image understanding and generation. Transfusion~\cite{zhou2024transfusion} and Show-o~\cite{xie2024show} integrate textual autoregressive modeling for image understanding and visual diffusion modeling for image generation.
Janus~\cite{wu2024janus} uses two different image representations, respectively for understanding and generation, to address the varying levels of information granularity required by the two tasks. 

However, previous methods either involve complex designs or face challenges such as computational cost and optimization stability. To address the issues, our method leverages Multimodal Mixture-of-Experts and a token folding strategy to construct a fully autoregressive MLLM, enabling synergistic high-resolution image understanding and generation. 
Experiments show that \modelname achieves state-of-the-art performance on various benchmarks.

\paragraph{Encoder-free MLLMs.}

Most existing MLLMs adopt an encoder-based framework that integrates a separate image encoder like CLIP~\cite{VLP:CLIP} into a pretrained LLM~\cite{TransF:Vicuna, qwen, cai2024internlm2}. 
Meanwhile, some recent attempts have also begun to develop encoder-free MLLMs architecture due to their simplicity.
Some works ~\cite{team2024chameleon, xie2024show, zhou2024transfusion, emu3} adopt VQ tokenizers~\cite{vqgan} to represent images as discrete tokens. Others~\cite{diao2024EVE, solo, mono_internvl} use simple linear projection (\emph{i.e.,} patch embedding layer) to embed the images. 
In this paper, we build an encoder-free MLLM using discrete image representation through VQ tokenizers, which has stronger reconstruction ability to support both understanding and generation. 

\paragraph{Token Folding and Unfolding.}

In language processing, early attempts like Funnel Transformer~\cite{funnel_transformer} and DataMUX~\cite{datamux} propose the downsample-upsample paradigm, \emph{i.e.} compress the token length in intermediate Transformer layers, to process long sequences efficiently.
MegaByte~\cite{megabyte} segments sequences into patches, and then uses a local sub-model within patches and a global model between patches.
HRED~\cite{hred} uses a lower-frequency model to process input sub-sequences without global context, and decodes outputs at the original data frequency.
Block Transformer~\cite{block_transformer} introduces a global-to-local structure to optimize the inference efficiency of autoregressive LLMs.
In this paper, we adopt the token folding and unfolding mechanism to support high-resolution image understanding and generation.
Since current visual tokenizers generate very long visual token sequences for high-resolution images, which is unsuitable to LLMs, we fold the visual token sequences before LLM modeling, and decode them back into the original local token sequences for image generation.

\section{\modelname}
\label{sec:method}

\subsection{Architecture}

\modelname is a unified MLLM with synergistic image understanding and generation capabilities. Fig.~\ref{fig:method} shows an overview of \modelname. Similar to previous work~\cite{lwm,team2024chameleon,emu3}, \modelname requires no externalized image generation models or additionally pretrained semantic encoders. It uses a single LLM with the unified next-token prediction objective for both tasks. Specifically, the input images and text are represented as discrete tokens by their corresponding tokenizers. The input multimodal token sequence consists of both image and text tokens, which always starts with a special token \verb|<s>| and ends with another special token \verb|</s>|. Special tokens \verb|<boi>| and \verb|<eoi>| are inserted before and after each image to indicate the beginning and end of the image, respectively. The multimodal token sequence is processed with a causal Transformer~\cite{TransF:Transformer} initialized from a pretrained LLM. The image and text output tokens are predicted autoregressively, then the image output tokens can be decoded into pixels with the pretrained VQ decoder.

\vspace{0.5em}
\noindent\textbf{Input Embedding with Visual Token Folding.} Existing discrete VQ-based image tokenizers require low feature downsample ratios to ensure reconstruction with fine details. This leads to long visual token sequences, limiting the use of high-resolution images in LLMs for detail-rich image understanding such as OCR-related tasks. To address this issue, we employ \emph{Token Folding} to increase the feature downsampling ratio within the MLLM. Specifically, given an image $I \in \mathbb{R}^{H\times W \times3}$ (\eg, $H=W=512$), an off-the-shelf pretrained discrete image tokenizer is used to encode the image into a 2D grid of discrete tokens with shape $h \times w$, where $h = H / p$ and $w = W / p$. Here, $p$ is the tokenizer's downsampling ratio (\eg, $p=8$). The visual token embeddings are then obtained from a learnable look-up table, and a learnable positional embedding $PE$ is added to each token embedding to preserve spatial prior. Similar to Pixel Shuffle, token embeddings are folded by concatenating every $m \times n$ token patch into one single visual token ($m=2$ and $n=8$ by default). Here, each folded token patch can be rectangular, following the latest practice of MLLMs for perception~\cite{Qwen2vl}. As shown in Fig.~\ref{fig:method} (a), this results in an additional downsampling ratio of $m \times n$, greatly compressing the token sequence for MLLM. For example, for an image of size $512\times512$, the original Emu3~\cite{emu3} tokenizer produces visual $4096$ tokens, while \modelname uses only $256$ tokens to represent it in MLLM with a token folding ratio of $2\times8$.

After Token Folding, an MLP is applied to each folded image patch embedding to align its feature dimension with the LLM's input dimension, yielding the final visual input features $x_V \in \mathbb{R}^{(\frac{h \cdot w}{m \cdot n}) \times d} $. The whole image embedding process can be formulated as:
\begin{equation}
    x_V = \operatorname{MLP}(\operatorname{TokenFold}(\operatorname{TokenEmbed}(I) + PE)).
\end{equation}

For text input, we employ the built-in word tokenizer and the text token embedding look-up table of the pretrained LLM to encode it into text embeddings $x_T$. 

The visual token embeddings $x_V$ are concatenated with the text token embeddings $x_T$ and learnable special token embeddings (\ie, \verb|<s>|, \verb|</s>|, \verb|<boi>|, \verb|<eoi>|), according to the input order to form the final multimodal inputs into the MLLM.

\vspace{0.5em}
\noindent\textbf{Incorporating Visual Capabilities with Multimodal Mixture-of-Experts (MMoEs).} To avoid substantial tuning of the pretrained LLM while incorporating visual capabilities into it, we introduce additional parameters to each LLM's Feed-Forward Network (FFN) layer as vision experts dedicated to image representation. Specifically, the FFN output of the $i$-th token is altered to
\begin{equation}
    \text{FFN-MMoE}(x_i) = \begin{cases}
\text{FFN}_V(x_i), &\text{if } x_i \text{ is visual}, \\
\text{FFN}_T(x_i), &\text{if } x_i \text{ is textual},
\end{cases}
\end{equation}
where $\text{FFN}_T$ denotes the original FFN in the pretrained LLM for text tokens, and $\text{FFN}_V$ denotes the vision expert FFN, which shares the same architecture as $\text{FFN}_T$ and is initialized from the corresponding pretrained text FFN.

Instead of tuning the entire pretrained LLM, we perform a two-stage alignment pretraining on the vision expert FFNs with mixed image understanding and generation data. By aligning the visual representations with the representation space of the pretrained LLM, we minimize the impact of the LLM's pretrained knowledge, ensuring the general perception and generalization capabilities. We introduce the two-stage alignment pretraining in Sec.~\ref{sec:method:training}.

\vspace{0.5em}
\noindent\textbf{Image Generation with Visual Token Unfolding.} As the token-folding operation reduces the number of visual tokens at the input side of MLLM, the visual tokens at the output must be unfolded to generate images.
We leverage a small causal transformer as the image generation head. Such head shares the same micro-architecture as the LLM but with fewer layers (\eg, 4) and has its own image token embedding and position embedding look-up tables, accompanied by an output classifier for VQ tokens. 
 
For the $i$-th folded image patch, $h_i$ is the corresponding output embedding generated by the MLLM. To predict the $j$-th discrete token id $v_{i}^{j}$ in this image patch, its probability distribution is formulated autoregressively as:
\begin{equation}
    p(v_{i}^{j}|v_{i}^{<j}, h_i) = \operatorname{Softmax}(f_{\theta}(v_{i}^{<j}, h_i)),
\end{equation}
where $f_{\theta}$ represents the causal Transformer head with parameters $\theta$, $v_{i}^{<j}$ denotes all generated VQ token ids before the $i$-th visual tokens in this image patch.
After generating VQ token ids for all image patches, we concatenate them to obtain the complete sequence of VQ token ids with the shape of $h \times w$ for image pixel decoding.

\subsection{Training}
\label{sec:method:training}

\textbf{Training Objective}.
The overall training objective of \modelname consists of two main components: text token prediction and image token prediction. Both modalities employ the same next-token prediction objective, formulated as
\begin{equation}
\label{equ:loss_overall}
\small
    \mathcal{L} = -\sum_{i \in \mathcal{T}}  \log p(\hat{x}_T^{i} = x_T^{i} | x^{<i}) - \lambda \sum_{i \in \mathcal{V}}  \log p(\hat{x}_V^{i} = x_V^{i} | x^{<i}),
\end{equation}
where $\mathcal{T}, \mathcal{V}$ are the index sets indicating all text and image tokens in the multimodal sequence, respectively, $\hat{x}_T^{i}$ and $\hat{x}_V^{i}$ denote the predicted text and image token at position $i$. The final loss objective is the weighted sum of the text and image losses, with a hyperparameter $\lambda$ to balance the relative loss weight between image understanding and image generation.

\begin{table}[t]
    \centering
    \resizebox{0.47\textwidth}{!}{
    \begin{tabular}{lccl}
    \hline
     & \textbf{Task} & \textbf{\#Sam.} & \textbf{Datasets} \\
    \hline
     & \multirow{2}{*}{Gen.} &  \multirow{2}{*}{667M}  & LAION-Aesthetics~\cite{Datasets:Laion-5b}, Megalith~\cite{megalith_10m}, SAM~\cite{TransF:SAM}, \\
        \cellcolor{white}&      &      & Objects365~\cite{shao2019objects365}, ImageNet-1k~\cite{deng2009imagenet}, \\
        \rowcolor{gray!15}
     \cellcolor{white}\multirow{-3}{*}{S.1}   & Und. & 667M & Laion-En~\cite{Datasets:Laion-5b}, COYO~\cite{kakaobrain2022coyo-700m}, SAM~\cite{TransF:SAM} \\
    \hline
    \cellcolor{white}& \multirow{3}{*}{Gen.} &  \multirow{3}{*}{170M} & LAION-Aesthetics~\cite{Datasets:Laion-5b}, Megalith~\cite{megalith_10m}, Objects365~\cite{shao2019objects365}, \\
        \cellcolor{white}&      &      & Unsplash~\cite{unsplash}, Dalle-3-HQ~\cite{dalle3}, JourneyDB~\cite{journeydb}, \\
        \cellcolor{white}&      &      & Internal Dataset \\
        \rowcolor{gray!15}
        \cellcolor{white}&  & \multirow{9}{*}{170M} & \textbf{Captioning:} Laion-En~\cite{Datasets:Laion-5b}, Laion-Zh~\cite{Datasets:Laion-5b}, COYO~\cite{kakaobrain2022coyo-700m}, \\
        \rowcolor{gray!15}
        \cellcolor{white}&      &      & GRIT~\cite{peng2023kosmos2}, COCO~\cite{Datasets:MSCOCO}, TextCaps~\cite{Datasets:TextCaps} \\
        \rowcolor{gray!15}
        \cellcolor{white}&      &      & \textbf{Detection:} Objects365~\cite{shao2019objects365}, GRIT~\cite{peng2023kosmos2}, All-Seeing~\cite{wang2023allseeing} \\
        \rowcolor{gray!15}
        \cellcolor{white}&      &      & \textbf{OCR (large):} Wukong-OCR~\cite{gu2022wukong}, LaionCOCO-OCR~\cite{Datasets:LAION-COCO}, \\
        \rowcolor{gray!15}
        \cellcolor{white}&      &      & Common Crawl PDF \\
        \rowcolor{gray!15}
        \cellcolor{white}&      &      & \textbf{OCR (small):} MMC-Inst~\cite{liu2023mmcinst}, LSVT~\cite{sun2019lsvt}, ST-VQA~\cite{biten2019stvqa}, \\
        \rowcolor{gray!15}
        \cellcolor{white}&      &      & RCTW-17~\cite{shi2017rctw17}, ReCTs~\cite{zhang2019rects}, ArT~\cite{chng2019art}, SynthDoG~\cite{kim2022synthdog}, \\
        \rowcolor{gray!15}
        \cellcolor{white}&      &      & ChartQA~\cite{Datasets:ChartQA}, CTW~\cite{yuan2019ctw}, DocVQA~\cite{Datasets:DocVQA}, TextOCR~\cite{singh2021textocr}, \\
        \rowcolor{gray!15}
    \cellcolor{white}\multirow{-12}{*}{S.2}  &  \multirow{-9}{*}{Und.}    &  \multirow{-9}{*}{170M} & COCO-Text~\cite{veit2016cocotext}, PlotQA~\cite{methani2020plotqa}, InfoVQA~\cite{mathew2022infographicvqa} \\
    \hline
    \end{tabular}
    }
    \caption{\textbf{Summary of datasets used in Visual Alignment Pretraining.} ``S.1'' and ``S.2'' denote the first and second stage. ``Gen.'' and ``Und.'' denote the image generation and understanding task. ``\#Sam.'' denotes the number of total samples seen during training of each task at each stage. Note that all data used for image understanding in the second stage is also used in InternVL-1.5~\cite{VLM:InternVL-1.5}.}
    \label{tab:datasets}
\end{table}

\begin{table*}[t]
    \centering
    \small
    \setlength{\tabcolsep}{4pt}
    \scalebox{0.91}{
    \begin{tabular}{lc|cccccccccc}
    \toprule
    \textbf{Model}      & \textbf{\#A-Param}     & \textbf{POPE}       & \textbf{MMB}             & \textbf{MMVet}      & \textbf{MMMU}        & \textbf{MME}  & \textbf{MME-P}       & \textbf{MathVista}              & \textbf{SEED-I}     & \textbf{OCRBench}        \\ 
    \midrule
    \midrule
    \multicolumn{11}{l}{\textbf{Understanding Only}} \\
    \midrule
    \rowcolor{Gray!5} \multicolumn{11}{l}{\small\textit{Encoder-based}}\\
    LLaVA-1.5~\cite{liu2023improved} & 7B & 85.9 & 64.3 & 31.1 & 35.4 & - & 1512 & - & 58.6 & - \\ 
    Mini-Gemini-2B~\cite{VLM:MiniGemini} & 3.5B & - & 59.8 & 31.1 & 31.7 & 1653 & - & 29.4 & - & - \\ 
    DeepSeek-VL-1.3B~\cite{lu2024deepseekvl} & 2B & 87.6 & 64.6 & 34.8 & 32.2 & 1532 & - & 31.1 & 66.7 & 409 \\ 
    PaliGemma-3B~\cite{beyer2024paligemma} & 2.9B & 87.0 & 71.0 & 33.1 & 34.9 & 1686 & - & 28.7 & 69.6 & 614 \\ 
    MiniCPM-V2~\cite{yao2024minicpm} & 2.8B & - & 69.1 & 41.0 & 38.2 & 1809 & - & 38.7 & 67.1 & 605 \\ 
    InternVL-1.5~\cite{VLM:InternVL-1.5} & 2B & - & 70.9 & 39.3 & 34.6 & 1902 & - & 41.1 & 69.8 & 654 \\ 
    Qwen2-VL~\cite{Qwen2vl} & 2B & - & 74.9 & 49.5 & 41.1 & 1872 & - & 43.0 & - & 809 \\ 
    \hline
    \rowcolor{Gray!5} \multicolumn{11}{l}{\small\textit{Encoder-free}}\\
    Fuyu-8B (HD)~\cite{VLM:Fuyu-8b} & 8B & - & 10.7 & 21.4 & - & - & - & - & - & -\\
    EVE-7B~\cite{diao2024EVE} & 7B & 83.6 & 49.5 & 25.6 & 32.3 & 1483 & - & 25.2 & 61.3 & 327 \\ 
    Mono-InternVL~\cite{mono_internvl} & 1.8B & - & 65.5 & 40.1 & 33.7 & 1875 & - & 45.7 & 67.4 & 767 \\
    \midrule
    \midrule
    \multicolumn{11}{l}{\textbf{Understanding \& Generation}} \\
    \midrule
    \rowcolor{Gray!5} \multicolumn{11}{l}{\small\textit{Encoder-based}}\\
    Emu~\cite{sun2023emu} & 14B & - & - & 36.3 & - & - & - & - & - & - \\ 
    Emu2~\cite{VLM:EMUv2} & 37B & - & 63.6 & 48.5 & 34.1 & - & 1345 & - & 62.8 & - \\ 
    SEED-X~\cite{seed_x} & 17B & 84.2 & 75.4 & - & 35.6 & - & 1436 & - & - & - \\ 
    LWM~\cite{lwm} & 7B & 75.2 & - & 9.6 & - & - & - & - & - & - \\ 
    DreamLLM~\cite{dong2023dreamllm} & 7B & - & 58.2 & 36.6 & - & - & - & - & - & - \\ 
    Janus~\cite{wu2024janus} & 1.3B & 87.0 & 69.4 & 34.3 & 30.5 & - & 1338 & - & 63.7 & - \\ 
    \hline
    \rowcolor{Gray!5} \multicolumn{11}{l}{\small\textit{Encoder-free}}\\
    Chameleon~\cite{team2024chameleon} & 7B & - & - & 8.3 & 22.4 & - & - & - & - & - \\ 
    Show-o~\cite{xie2024show} & 1.3B & 84.5 & - & - & 27.4 & - & 1233 & - & - & -\\     
    VILA-U~\cite{wu2024vila} & 7B & 85.8 & - & 33.5 & - & - & 1402 & - & 59.0 & - \\ 
    Emu3-Chat~\cite{emu3} & 8B & 85.2 & 58.5 & 37.2 & 31.6 & - & - & - & 68.2 & 687 \\ 
    \rowcolor{Gray!15} \modelname (Ours) & 2.4B & 85.3 & 53.7 & 34.5 & 34.2 & 1837 & 1381 & 42.7 & 62.0 & 721\vspace{-0.6mm}\\

    \bottomrule
    \end{tabular}
    }
    \caption{\textbf{Results on general MLLM benchmarks.} Our model with 2.4B parameters achieves competitive image understanding performance compared with significantly larger encoder-free unified MLLMs such as Emu3-Chat-8B~\cite{emu3}.
    }   
    \vspace{-0.5em}
    \label{tab:mllm_benchmark}
\end{table*}

\vspace{0.5em}
\noindent{\textbf{Visual Alignment Pretraining}}.
To preserve the LLM's pretrained knowledge, we conduct a progressive two-stage alignment pretraining strategy, with both stages utilizing a mixture of data for image understanding and generation. The detailed dataset composition is shown in Tab.~\ref{tab:datasets}. 

The first stage aims to bridge visual elements with concepts in the representation space of the pretrained LLM, thereby obtaining basic semantic understanding and image generation abilities. To avoid interfering with the LLM's pretrained knowledge, we freeze the parameters of the LLM components and only train the image-specific parameters (\ie, the visual token embedding and projection layers, the vision experts in MLLM, and the visual token unfolding head). For image understanding, we use the large-scale noisy image-text pair data LAION-En~\cite{Datasets:Laion-5b} and Coyo-700M~\cite{kakaobrain2022coyo-700m} for basic concept learning, while incorporating a portion of synthesized captions from samples in \cite{Datasets:Laion-5b,kakaobrain2022coyo-700m,TransF:SAM} generated by InternVL-7B to achieve better semantic alignments. For image generation, apart from the large-scale noisy LAION-Aesthetics data~\cite{Datasets:Laion-5b}, we follow~\cite{pixart, wu2024janus} to accelerate the pixel dependency learning with~\cite{deng2009imagenet} and improve the learning of object concepts and relations through~\cite{TransF:SAM, megalith_10m, shao2019objects365}. To distinguish the image understanding and generation tasks, we use the prompt of ``Provide a one-sentence caption for the image'' for understanding data, while adding ``Generate an image of:'' before the text prompts of the generation data. 

In the second stage, we further integrate visual capabilities into the pretrained LLM by training the image-specific parameters and the self-attention layers with high-quality mixed data. Specifically, for image understanding, we follow~\cite{mono_internvl} to sample from the high-quality pretraining data of InternVL-1.5~\cite{VLM:InternVL-1.5}, resulting in 170 million samples with task-related prompts. For image generation, we select the data with high aesthetic scores and caption quality, resulting in 20 million samples from~\cite{megalith_10m, unsplash, shao2019objects365, dalle3, journeydb} as well as 5 million high-quality internal data.

Throughout both stages, \modelname is trained simultaneously for image understanding and generation. To enhance the image understanding capability and take advantage of \modelname's ability to process high-resolution images, we implement a dynamic resolution strategy for understanding tasks following InternVL-1.5~\cite{VLM:InternVL-1.5} in the second stage and set the maximum number of image tiles to 6.

\vspace{0.5em}
\noindent\textbf{Joint Instruction Tuning}.
During the instruction tuning stage, we unfreeze all the model parameters. For image understanding, we adopt the dataset from InternVL-1.5, including around 5M bilingual instructions for supervised learning, covering various tasks such as visual question answering, multimodal dialogue, mathematics, knowledge, etc. We also increase the maximum number of image tiles to 12 to handle high-resolution images. For image generation, we solely use the 10M internal dataset to further enhance the image generation quality.

\section{Experiments}
\label{sec:experiment}

\begin{table*}[h!]
    \centering
    \small
    \setlength{\tabcolsep}{7pt}
    \resizebox{0.88\linewidth}{!}{
    \begin{tabular}{lc|ccccccccc}
    \toprule
    \textbf{Method}      &\textbf{\#A-Param}        &\textbf{TextVQA}        &\textbf{SQA-I}          &\textbf{GQA}       &\textbf{DocVQA}            &\textbf{AI2D}        &\textbf{ChartQA} 
     &\textbf{InfoVQA} \\ 
    \midrule
    \midrule
    \multicolumn{9}{l}{\textbf{Understanding Only}} \\
    \midrule
    \rowcolor{Gray!5} \multicolumn{9}{l}{\small\textit{Encoder-based}}\\
    MobileVLM-V2~\cite{chu2024mobilevlm} & 1.7B & 52.1 & 66.7 & 59.3 & - & - & - & - \\
    Mini-Gemini-2B~\cite{li2024miniGemini} & 3.5B & 56.2 & - & - & 34.2 & - & - & - \\
    PaliGemma-3B~\cite{beyer2024paligemma} & 2.9B & & 68.1 & - & - & - & 68.3 & - \\
    MiniCPM-V2~\cite{yao2024minicpm} & 2.8B & 74.1 & - & - & 71.9 & - & - & - \\
    InternVL-1.5~\cite{VLM:InternVL-1.5} & 2B & 70.5 & 84.9 & 61.6 & 85.0 & 69.8 & 74.8 & 55.4 \\
    \hline
    \rowcolor{Gray!5} \multicolumn{9}{l}{\small\textit{Encoder-free}} \\
    EVE-7B~\cite{diao2024EVE} & 7B & 51.9 & 63.0 & 60.8 & - & - & - & - \\
    Mono-InternVL~\cite{mono_internvl} & 1.8B & 72.6 & 93.6 & 59.5 & 80.0 & 68.6 & 73.7 & 43.0 \\
    \midrule
    \midrule
    \multicolumn{9}{l}{\textbf{Understanding \& Generation}} \\
    \hline
    \rowcolor{Gray!5} \multicolumn{9}{l}{\small\textit{Encoder-based}} \\
    Emu2~\cite{VLM:EMUv2} & 37B & 66.6 & - & 65.1 & - & - & - & - \\
    LWM~\cite{lwm} & 7B & 18.8 & 47.7 & 44.8 & - & - & - & - \\
    DreamLLM~\cite{dong2023dreamllm} & 7B & 41.8 & - & - & - & - & - & - \\
    MM-Interleaved~\cite{tian2024mminterleaved} & 13B & 61.0 & - & 60.5 & - & - & - & - \\
    Janus~\cite{wu2024janus} & 1.3B & - & - & 59.1 & - & - & - & - \\
    \hline
    \rowcolor{Gray!5} \multicolumn{9}{l}{\small\textit{Encoder-free}}\\
    Chameleon$^\diamond$~\cite{team2024chameleon} & 7B & 4.8 & 47.2 & - & 1.5 & 46.0 & 2.9 & 5.0   \\
    Show-o~\cite{xie2024show} & 1.3B & - & - & 61.0 & -  & - & - & - \\
    VILA-U~\cite{wu2024vila} & 7B & 60.8 & - & 60.8 & - & - & - & -  \\
    Emu3-Chat~\cite{emu3} & 8B & 64.7 & 89.2 & 60.3 & 76.3 & 70.0 & 68.6 & 43.8 \\
    \rowcolor{Gray!15} 
    \modelname (Ours) & 2.4B & 67.5 & 92.6 & 59.7 & 76.6 & 60.8 & 73.4 & 37.5\vspace{-0.6mm} \\
    \bottomrule
    \end{tabular}
    }
    \caption{\textbf{Comparison with existing MLLMs on visual question answering benchmarks.} \#A-Params denotes the number of activated parameters during inference. $^\diamond$Some results of Chameleon are sourced from \cite{mono_internvl}.}
    \label{tab:vqa_benchmark}
\end{table*}

\subsection{Implementation Details}

\modelname is built upon InternLM2-1.8B~\cite{cai2024internlm2}, using the same text tokenizer and conversation format. The discrete image tokenizer originates from Emu3~\cite{emu3}, characterized by a codebook size of 32,768 and a spatial downsampling rate of 8. The input image is resized to $512\times 512$.
For image generation data, the short edge of the image is resized to 512 and the long edge is cropped to 512.
The total number of model parameters is 3.6B, of which the number of activation parameters is 2.4B. 
In the pretraining phase, the global batch size for image understanding and generation tasks is 6988 for stage 1 and 5090 for stage 2, respectively.
The loss weight hyperparameter $\lambda$ is set to 2.
The instruction tuning phase is trained for 3 epochs in total.
Following previous works~\cite{llamagen, wu2024janus}, classifier-free guidance (CFG) strategy is also implemented for image generation. During training, we randomly replace the original user caption prompt with ``Here is a random image \verb|<UNCOND>|:'' with a probability of 10\%, where \verb|<UNCOND>| is a learnable special token embedding.
During inference, the logit of each unfolded image token is calculated as: $l_g = l_u + s(l_c - l_u)$, where $l_c, l_u$ are the conditional and unconditional logits, respectively. $s$ is the CFG-scale with default number of 7.5.

Due to the limited space, please refer to the supplementary material for more detailed training configurations.

\subsection{Image Understanding}

\noindent\textbf{Evaluation Benchmarks.} To evaluate the general multimodal understanding capabilities of \modelname, we compare with image understanding models as well as unified image understanding and generation models on 8 comprehensive multimodal benchmarks including MMBench-EN \emph{test}~\cite{Datasets:MMBench}, MMVet~\cite{Datasets:MM-vet}, MMMU \emph{val}~\cite{Datasets:MMMU}, MME~\cite{Datasets:MME}, MathVista \emph{test-mini}~\cite{Datasets:Mathvista}, POPE~\cite{Datasets:POPE}, SEED-Image~\cite{Datasets:Seed-bench}, and OCRBench~\cite{liu2023ocrbench}. These general benchmarks covers assessment of various capabilities for visual question answering, document and chart interpretation, and other complex visual scenarios. We further evaluate model's VQA performances on 7 widely-adopted benchmarks including TextVQA \emph{val}~\cite{Datasets:TextVQA}, ScienceQA \emph{test}~\cite{Datasets:ScienceQA}, GQA \emph{test-dev}~\cite{Datasets:GQA}, DocVQA \emph{test}~\cite{Datasets:DocVQA}, AI2D \emph{test}~\cite{Datasets:AI2D}, ChartQA \emph{test}~\cite{Datasets:ChartQA}, and InfographicsVQA \emph{test}~\cite{mathew2022infographicvqa}. Part of the results are evaluated using VLMEvalKit~\cite{duan2024vlmevalkit} or sourced from the OpenCompass leaderboard~\cite{opencompass2023}.

\vspace{0.5em}
\noindent\textbf{Results}. Evaluation results are shown in Tab.~\ref{tab:mllm_benchmark} and Tab.~\ref{tab:vqa_benchmark}.
Compared with existing encoder-free unified MLLMs, our \modelname with 2.4B parameters surpasses previous methods (especially for encoder-free unified MLLMs) with comparable parameter sizes while achieving comparable performance to models with significantly larger parameter sizes, showcasting its competitive image understanding capability. Notably, on image understanding benchmarks requiring high-resolution detailed image comprehension, such as OCRBench, TextVQA, DocVQA, and ChartQA, our \modelname achieves results superior to much larger encoder-free MLLMs such as Emu3-Chat-8B~\cite{emu3}, highlighting its advantages with high-resolution image processing capabilities. Moreover, as a encoder-free unified MLLM, \modelname also obtains image understanding performance competitive to encoder-based understanding-only MLLMs such as LLaVA-1.5~\cite{VLM:LLaVA-1.5}, while surpassing larger encoder-free task-specific MLLMs such as EVE-7B~\cite{diao2024EVE} and Fuyu-8B (HD)~\cite{VLM:Fuyu-8b}, demonstrating its great potential of unifying image understanding and generation.

\begin{table*}[ht]
    \centering
    \small
    \setlength{\tabcolsep}{5pt}
    \resizebox{0.9\linewidth}{!}{
    \begin{tabular}{lc|cccccc|c}
    \toprule
    \textbf{Method} & \textbf{\# A-Param} & \textbf{Single Obj.} & \textbf{Two Obj.} & \textbf{Counting} & \textbf{Colors} & \textbf{Position} & \textbf{Color Attri.} & \textbf{Overall$\uparrow$} \\
    \hline
    \midrule
    \multicolumn{8}{l}{\textbf{Generation Only}} \\
    \hline
    LlamaGen~\cite{llamagen} & 0.8B & 0.71 & 0.34 & 0.21 & 0.58 & 0.07 & 0.04 & 0.32 \\
    LDM~\cite{rombach2021high} & 1.4B & 0.92 & 0.29 & 0.23 & 0.70 & 0.02 & 0.05 & 0.37 \\
    SDv1.5~\cite{rombach2021high} & 0.9B & 0.97 & 0.38 & 0.35 & 0.76 & 0.04 & 0.06 & 0.43 \\
    SDXL~\cite{podell2023sdxl} & 2.6B & 0.98 & 0.74 & 0.39 & 0.85 & 0.15 & 0.23 & 0.55 \\
    PixArt-$\alpha$~\cite{pixart} & 0.6B & 0.98 & 0.50 & 0.44 & 0.80 & 0.08 & 0.07 & 0.48 \\
    DALL-E 2~\cite{ramesh2022hierarchical} & 6.5B & 0.94 & 0.66 & 0.49 & 0.77 & 0.10 & 0.19 & 0.52 \\
    \hline
    \midrule
    \multicolumn{8}{l}{\textbf{Understanding \& Generation}} \\
    \hline
    SEED-X$\dag$~\cite{seed_x} & 17B & 0.97 & 0.58 & 0.26 & 0.80 & 0.19 & 0.14 & 0.49 \\
    Show-o~\cite{xie2024show} & 1.3B & 0.95 & 0.52 & 0.49 & 0.82 & 0.11 & 0.28 & 0.53 \\
    LWM~\cite{lwm} & 7B & 0.93 & 0.41 & 0.46 & 0.79 & 0.09 & 0.15 & 0.47 \\
    Chameleon~\cite{team2024chameleon} & 34B & - & - & - & - & - & - & 0.39 \\
    Emu3-Gen~\cite{emu3} & 8B & 0.98 & 0.71 & 0.34 & 0.81 & 0.17 & 0.21 & 0.54 \\
    Janus~\cite{wu2024janus} & 1.3B & 0.97 & 0.68 & 0.30 & 0.84 & 0.46 & 0.42 & 0.61 \\
    \rowcolor{gray!15}
    \modelname (Ours) & 2.4B & 0.99 & 0.71 & 0.34 & 0.87 & 0.37 & 0.37 & 0.61\vspace{-0.3mm} \\
    \hline
    \end{tabular}
    }
    \caption{\textbf{Evaluation of text-to-image generation on GenEval~\cite{ghosh2024geneval} benchmark.} \#A-Params denotes the number of activated parameters during inference. $\dag$ indicates models with external pretrained diffusion model. Obj.: Object. Attri.: Attribution.}
\end{table*}

\begin{table}[t]
    \centering
    \small
    \setlength{\tabcolsep}{3pt}
    \resizebox{0.42\textwidth}{!}{
    \begin{tabular}{lc|cc}
        \toprule
        \textbf{Model} & \textbf{\#A-Param} & \textbf{MS-COCO$\downarrow$} & \textbf{MJHQ$\downarrow$} \\
        \midrule
        \midrule
        \multicolumn{4}{l}{\textbf{Generation Only}} \\
        \midrule
         DALL-E~\cite{ramesh2021zero}  & 12B & 27.50 & - \\
         LDM~\cite{rombach2021high} & 1.4B & 12.64 & - \\
         GLIDE~\cite{nichol2021glide} & 5B & 12.24 & - \\
         DALL-E 2~\cite{ramesh2022hierarchical}  & 6.5B & 10.39 & - \\
         RAPHAEL~\cite{xue2024raphael} & 3B & 6.61 & - \\
         Imagen~\cite{saharia2022photorealistic} & 34B & 7.27 & - \\
         SDv1.5~\cite{rombach2021high}  & 0.9B & 9.62 & - \\
         SDXL~\cite{podell2023sdxl}  & 0.9B & 7.38 & 8.76 \\
         PixArt-$\alpha$~\cite{pixart}  & 0.6B & 7.32 & 6.14 \\
        \midrule
        \midrule
        \multicolumn{4}{l}{\textbf{Understanding \& Generation}} \\
        \midrule
         NExT-GPT~\cite{VLM:NeXTGPT}  & 13B & 11.18 & - \\
         SEED-X~\cite{seed_x}  & 17B & 14.99 & - \\
         Show-o~\cite{xie2024show}  & 1.3B & 9.24 & 15.18 \\
         LWM~\cite{lwm} & 7B & 12.68 & 17.77 \\
         VILA-U~\cite{wu2024vila} & 7B & - & 7.69 \\
         Emu3-Gen~\cite{emu3} & 8B & 19.3 & - \\
         Janus~\cite{wu2024janus} & 1.3B & 8.53 & 10.10 \\
         \rowcolor{gray!15}
         \modelname (Ours) & 2.4B & 7.65 & 6.10\vspace{-0.7mm}\\
        \bottomrule
    \end{tabular}
    }
    \caption{\textbf{Image generation results on MSCOCO-30K~\cite{Datasets:MSCOCO} and MJHQ-30K~\cite{Datasets:MJHQ} datasets.} FID~\cite{heusel2017gansfid} is reported. \#A-Param denotes the number of activated parameters during inference.}
    \label{tab:benchmark_coco}
\end{table}

\subsection{Image Generation}

\noindent\textbf{Evaluation Benchmarks}. We use the MSCOCO-30K~\cite{Datasets:MSCOCO}, MJHQ-30K~\cite{Datasets:MJHQ}, and GenEval~\cite{ghosh2024geneval} benchmarks to evaluate our model's image generation capabilities. For MSCOCO-30K and MJHQ-30K, we generate 30k images and compare them with the reference images and use Fr\'{e}chet Inception Distance (FID)~\cite{heusel2017gansfid} to assess the overall generation quality. For GenEval, we generate four images for each prompt and utilize its official framework to assess our model's object-level image-text alignment.

\vspace{0.5em}
\noindent\textbf{Results on MSCOCO and MJHQ.} Tab.~\ref{tab:benchmark_coco} shows the zero-shot FID of our model on MSCOCO 30K~\cite{Datasets:MSCOCO}. Compared with previous generation-only models such as GLIDE~\cite{nichol2021glide} and DALL-E 2\cite{ramesh2022hierarchical}, our method can achieve better FID scores. Compared with previous unified MLLMs for both image understanding and generation, \modelname can achieve competitive performance without using an external diffusion model. In particular, compared with Emu3~\cite{emu3} that use the same tokenizer, our method has a significant improvement in FID scores with less model parameters. We believe this is because the usage of vision experts simplifies the training difficulty. We also evaluate our model's ability to generate high-quality aesthetic images on MJHQ~\cite{Datasets:MJHQ}, as shown in the Tab.~\ref{tab:benchmark_coco}. Compared with previous generation-only methods, \modelname achieves competitive generation performance. These results validate that our method applies to both natural images and synthetic aesthetic images.

\vspace{0.5em}
\noindent\textbf{Results on GenEval.} Following previous studies, we evaluate our model's text-to-image generation capabilities on the GenEval benchmark~\cite{ghosh2024geneval} from six dimensions: ``single object'', ``two objects'', ``number'', ``color'', ``position'', and ``color attribution''. Our model achieve competitive overall scores with previous generation-only models of similar sizes. \modelname performs comparably to Janus~\cite{wu2024janus}, which uses independent encoders for perception and generation, demonstrating the effectiveness of using vision experts in our approach. Compared to Emu3~\cite{emu3}, our model achieves better overall performance with fewer parameters.

\section{Ablation Study}

In this section, we ablate the effectiveness of the two important techniques of \modelname, \ie, \emph{token folding} and \emph{progressive alignment pre-training with MMoEs}. In this ablation study, we use Qwen2-0.5B-Instruct~\cite{yang2024qwen2} as the initialized LLM and image size 256 unless otherwise specified.

\subsection{Effectiveness of Token Folding}

To verify the effectiveness of token folding on high-resolution image understanding, we compare \modelname with the baseline version without token folding and the dynamic resolution strategy on image understanding tasks. Specifically, the baseline model directly use the tokenized sequence as the input image sequence without token folding, where the input image size is $256\times256$ and the tokenized sequence length is 1024. Meanwhile, for the model with token folding, we follow InternVL-1.5~\cite{VLM:InternVL-1.5} to implement the dynamic resolution strategy to provide high-resolution input images. For fair comparison, we use a token folding ratio of $2\times4$ and control the maximum number of dynamic image patches so that the average length of image token sequence after token folding is also 1024.

We train the models with a subset of stage 2 (S.2) understanding data, and evaluate the pre-trained models on VQA benchmarks. Results are shown in Tab.~\ref{tab:benchmark_tokenfold}. On datasets requiring precise understanding of detailed image information such as TextVQA, DocVQA, ChartVQA, and InfographicVQA, the model with token folding achieves significantly better results, demonstrating its advantages of high-resolution image understanding. 

\begin{table}[hbt]
    \centering
    \small
    \setlength{\tabcolsep}{3pt}
    \resizebox{0.47\textwidth}{!}{
    \begin{tabular}{l|cccccc}
        \toprule
        \textbf{Model} & \textbf{TextVQA} & \textbf{GQA} & \textbf{DocVQA} & \textbf{AI2D} & \textbf{ChartQA} & \textbf{InfoVQA} \\
        \midrule
        \emph{w/o} token folding & 18.7 & 45.3 & 14.7 & 42.0 & 20.9 & 18.7 \\
        \rowcolor{gray!15}
        \emph{w/} token folding & 35.0 & 45.1 & 36.7 & 42.1 & 49.7 & 21.1 \\

        \bottomrule
    \end{tabular}
    }
    \caption{\textbf{Comparison between models with and without token-folding on VQA benchmarks.} The model with token folding demonstrates significant performance improvements with the same image token sequence length. }
    \label{tab:benchmark_tokenfold}
\end{table}

\subsection{Effectiveness of the Progressive Alignment Pretraining with MMoEs}

\begin{table*}[hbt]
    \centering
    \small
    \setlength{\tabcolsep}{2pt}
    \resizebox{0.98\textwidth}{!}{
    \begin{tabular}{ll|cccccc|cccc|c}
        \toprule
        \multirow{2}{*}{\textbf{Stage}} & \multirow{2}{*}{\textbf{Strategy}} & \multicolumn{6}{c|}{\textbf{VQA Benchmarks} $\uparrow$} & \multicolumn{4}{c|}{\textbf{NLP Benchmarks} $\uparrow$} & \multicolumn{1}{c}{\textbf{T2I Benchmark} $\downarrow$ } \\
         & & \textbf{TextVQA} & \textbf{GQA} & \textbf{DocVQA} & \textbf{AI2D} & \textbf{ChartQA} & \textbf{InfoVQA} & \textbf{MMLU} & \textbf{CMMLU} & \textbf{AGIEVAL} & \textbf{MATH} & \textbf{MSCOCO} \\
        \midrule
        \multicolumn{2}{l|}{\color{gray} Baseline (Qwen2-0.5B)} & - & - & - & - & - & - & \color{gray} 42.3 & \color{gray} 51.4 & \color{gray} 29.3 & \color{gray} 12.1 & - \\
        S.1 + S.2 & Full & 14.3 & 42.9 & 11.3 & 24.7 & 12.4 & 12.6 & 23.1 & 23.0 & 8.1 & 0.9 & 30.7 \\
        S.1 only & Progressive & 0.1 & 13.0 & 0.2 & 0.3 & 0.0 & 0.0 & 42.3 & 51.4 & 29.3 & 12.1 & 28.3 \\
        S.2 only & Progressive & 8.7 & 36.9 & 8.6 & 40.9 & 11.7 & 16.2 & 37.6 & 45.3 & 28.9 & 7.2 & 34.9 \\
        \rowcolor{gray!15}
        S.1 + S.2 & Progressive &  13.2 & 41.2 & 11.4 & 41.9 & 12.8 & 17.0 & 39.3 & 48.2 & 26.2 & 8.9 & 20.2\vspace{-0.6mm} \\
        \bottomrule
    \end{tabular}
    }
    \caption{\textbf{Zero-shot performance of different pre-training strategies.} ``S.1'' and ``S.2'' denote the first and second pre-training stage. ``Full'' and ``Progressive'' denote the full parameter tuning and our progressive tuning strategy with MMoEs, respectively. FID~\cite{heusel2017gansfid} is reported for text-to-image generation (T2I) on MSCOCO~\cite{Datasets:MSCOCO}.}
    \label{tab:benchmark_pretrain}
\end{table*}

We ablate our proposed visual alignment pre-training strategy on various benchmarks, including visual question answering (VQA), natural language processing (NLP) and text-to-image (T2I) generation, as shown in Tab.~\ref{tab:benchmark_pretrain}. To ensure fair comparison, neither token folding nor dynamic resolution strategies are employed. For experimental efficiency, only 1/6 of the training data is used for both stages.

The results show that our progressive strategy matches or exceeds the fully parameter-trained strategy on VQA benchmarks and significantly outperforms it on text-to-image generation benchmarks. Meanwhile, on NLP benchmarks, our model with progressive alignment pre-training delivers results much closer to the pre-trained LLM (Qwen2-0.5B-Instruct) compared with the fully parameter-trained model. This validates that our approach effectively preserves the original knowledge in the pre-trained LLM while learning robust visual representations.
Furthermore, the two-stage training strategy outperforms training solely with stage 1 or stage 2, particularly on VQA and text-to-image generation benchmarks. This underscores the importance of learning basic visual concepts and pixel dependencies from large-scale noisy data, as well as enhancing image-text alignment and image aesthetics with high-quality data.

\subsection{Analysis of Relationship Between Image Generation and Understanding}

We provide visualization and analysis to understand the relationship between image generation and understanding tasks, \ie how the two tasks might be related in terms of their processing or feature utilization.

\begin{figure}[ht]
  \centering
  \includegraphics[width=0.9\linewidth]{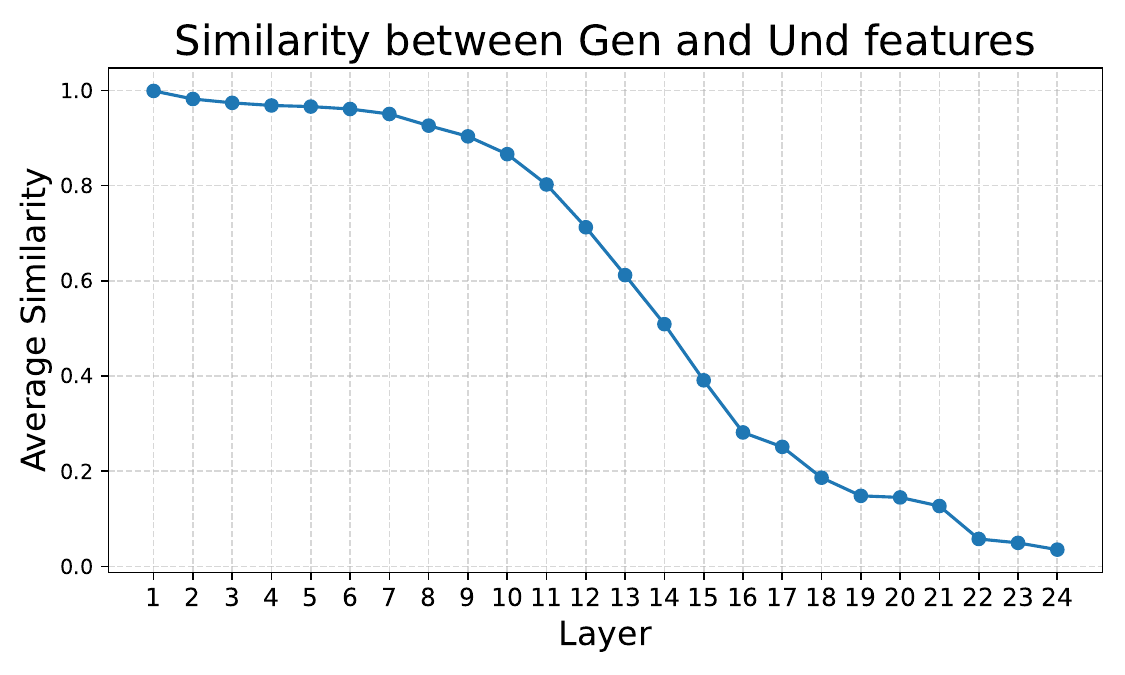}
   \caption{\textbf{Cosine similarity of visual features between generation and understanding tasks across different layers}. The representations of the image understanding and generation tasks are similar in shallow layers but disentagle in deeper layers.
   } 
   \label{fig:similarity}
\end{figure}

\noindent\textbf{Image Feature Similarity.}~We first analyze whether the two tasks share similar representations. We use the same input image paired with text instructions of generation or understanding, and compute the cosine similarity between visual features of the two tasks at each layer. As shown in Fig.~\ref{fig:similarity}, the two features are nearly identical (0.999) at shallower layers, but the similarity decreases as layers deepen. It finally reaches a near-zero value (0.035) at the last layer, suggesting that the two representations are disentangled. This observation implies that while image generation and understanding may share foundational visual representations in the early stages, they develop task-specific representations based on different instructions of image generation and understanding at deeper layers.

\begin{figure*}[!h]
  \centering
  \includegraphics[width=0.99\linewidth]{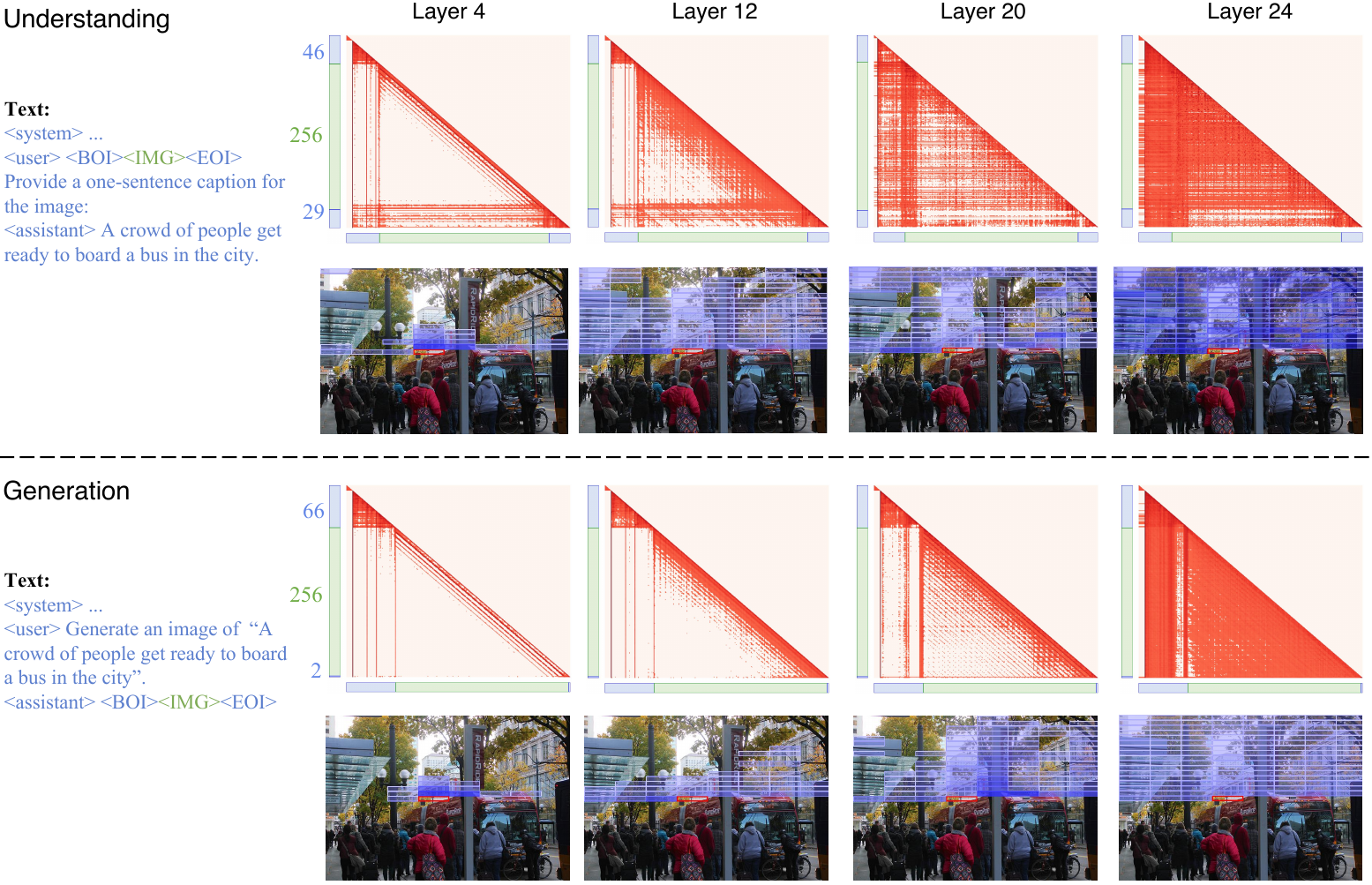}
   \caption{\textbf{Attention map visualization of understanding and generation tasks.} In the second and fourth rows, we visualize a query token \textcolor{red}{(red)} and its attended tokens \textcolor{blue}{(blue)} in the input image. Each token corresponds to a horizontal rectangular area in the original image due to the $2 \times 4$ token folding. Darker blue indicates larger attention weights.
   } 
   \label{fig:attn_map}
\end{figure*}

\vspace{0.5em}
\noindent\textbf{Attention Map Visualization.}~In Fig.~\ref{fig:attn_map}, we further investigate whether the two tasks have similar attention map patterns. We discover that in both tasks, locality is present at early layers, where visual tokens only attend to its nearby tokens (\ie near the diagonal). Text tokens and images have few interactions with each other. As layers deepen, longer dependency is observed, and finally global interactions are achieved at the last layer. Text and image also interact more often than at shallower layers. The attention weight also displays a periodicity nature, such as in Layer 4. Visualization in the input image suggests that the period is the number of tokens in each row, validating the locality. When comparing the attention maps in the two tasks, we observe that locality is more obvious in generation than in understanding at the same layer. This can be explained that local details are required to generate a spatially consistent and semantically coherent image, while understanding the whole image requires global context.

\section{Conclusion}
\label{sec:conclusion}

In this paper, we introduce \modelname, an encoder-free MLLM that effectively unifies image understanding and generation within a simplified framework. By leveraging token folding and vision experts, \modelname addresses the complexities of high-resolution image processing while maintaining the integrity of pretrained language model knowledge. Our approach eliminates dependencies on external diffusion models or additional semantic encoder pretraining, achieving competitive performance across various benchmarks with a relatively small parameter size. The experiment results underscore \modelname's potential as a scalable and efficient solution for future unified MLLMs.

\paragraph{Acknowledgments~}
This work is supported by the National Key R\&D Program of China (NO. 2022ZD0161300), by the National Natural Science Foundation of China (62376134).

\clearpage
\newpage
{
    \small
    \bibliographystyle{ieeenat_fullname}
    \bibliography{main}
}

\clearpage

\appendix
\onecolumn

\section{Detailed Training Configurations}

More detailed hyper-parameters used in the training stages are listed in Tab.~\ref{tab:hyperparam}.

\begin{table}[h]
    \centering
    \small
    \vspace{-0.0em}
    \resizebox{0.7\linewidth}{!}{
    \begin{tabular}{l|cc|c}
         \toprule
         \multirow{2}{*}{Configuration}     & \multicolumn{2}{c|}{Alignment Pre-training} & Instruction \\
         & S.1 & S.2 & Tuning \\
         \midrule
                  Maximum number of image tiles   & $1$ & $6$ & $12$ \\
                           LLM sequence length      & $4,096$ & $8,192$& $16,384$\\
         Use thumbnail            & \ding{55} & \ding{51} & \ding{51} \\
         Global batch size (per-task)        & $6,988$ & $5,090$ & $1,760$\\
         Peak learning rate       & $1e^{-4}$ & $5e^{-5}$ & $5e^{-5}$\\
         Learning rate schedule   & constant with warm-up & cosine decay & cosine decay\\
         Weight decay             & $0.05$ & $0.05$ & $0.01$ \\
         Training steps           & $95$k & $35$k & $12$k\\
         Warm-up steps            & \multicolumn{3}{c}{$200$} \\
         Optimizer                & \multicolumn{3}{c}{AdamW} \\
         Optimizer hyperparameters & \multicolumn{3}{c}{$\beta_{1}=0.9, \beta_{2}=0.95, eps=1e^{-8}$} \\
         Gradient accumulation    & \multicolumn{3}{c}{$1$} \\
         Numerical precision      & \multicolumn{3}{c}{$\mathtt{bfloat16}$} \\
         \bottomrule
    \end{tabular}
    }
    \caption{\textbf{Hyper-parameters used in the alignment pre-training and instruction tuning stages.}}
    \label{tab:hyperparam}
\end{table}

\section{Visualization}

For qualitative evaluation, we visualize examples for image understanding and image generation as follows.

\begin{figure*}[ht]
  \centering
  \includegraphics[width=0.98\textwidth]{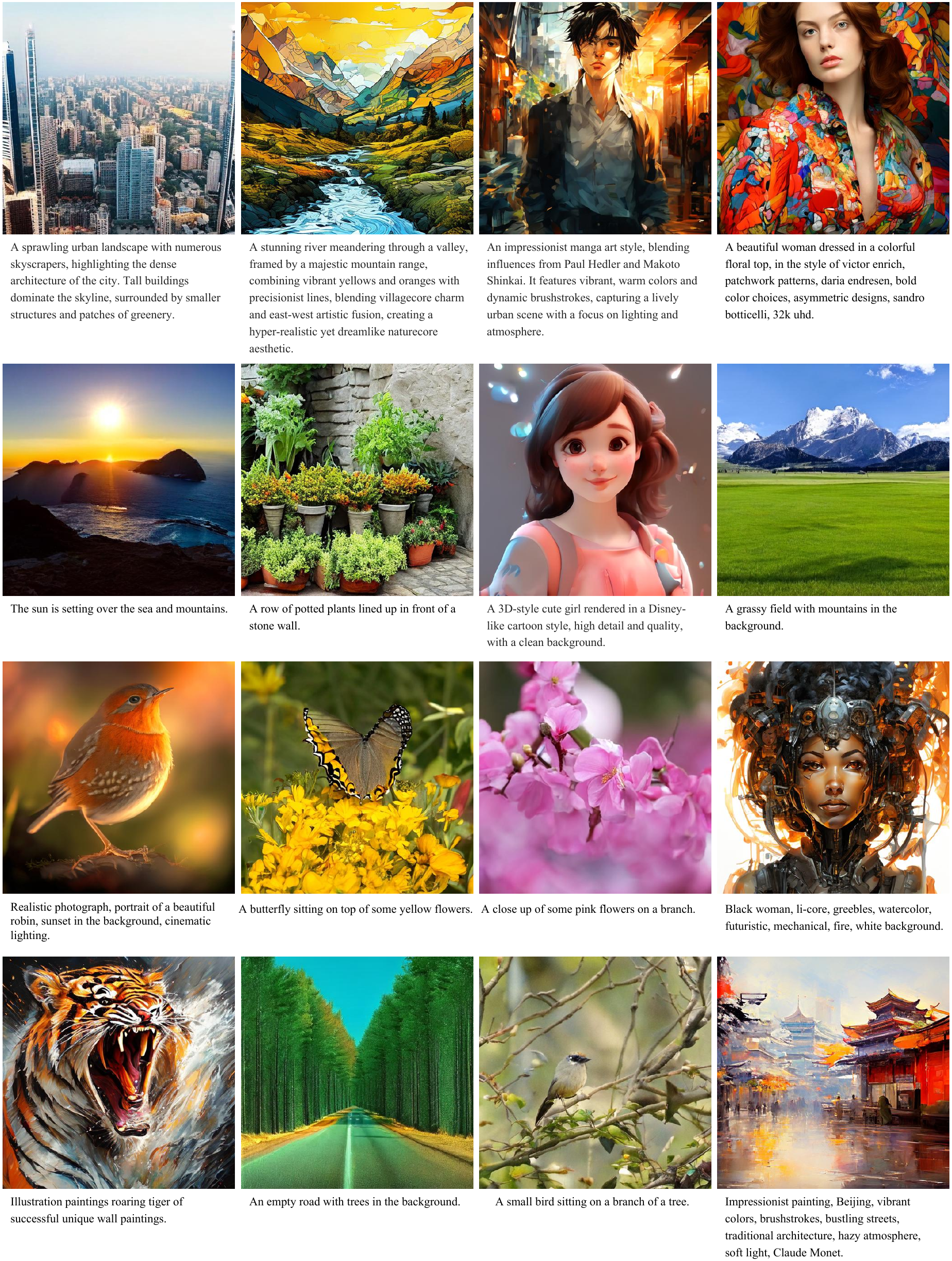}
  \vspace{-0.5em}
   \caption{\textbf{Qualitative results of image generation.} The images are of size $512 \times 512$.} 
   \label{fig:t2i_vis_natural}
\end{figure*}

\begin{figure*}[ht]
\begin{flushleft}\textbf{Image Captioning}\end{flushleft}
    \centering
\begin{tcolorbox}[rounded corners, colback=white, colframe=black, left=5pt, right=5pt, top=5pt, bottom=5pt]
    \begin{minipage}[ft]{0.97\textwidth} 
    \centering
        \includegraphics[width=0.6\linewidth]{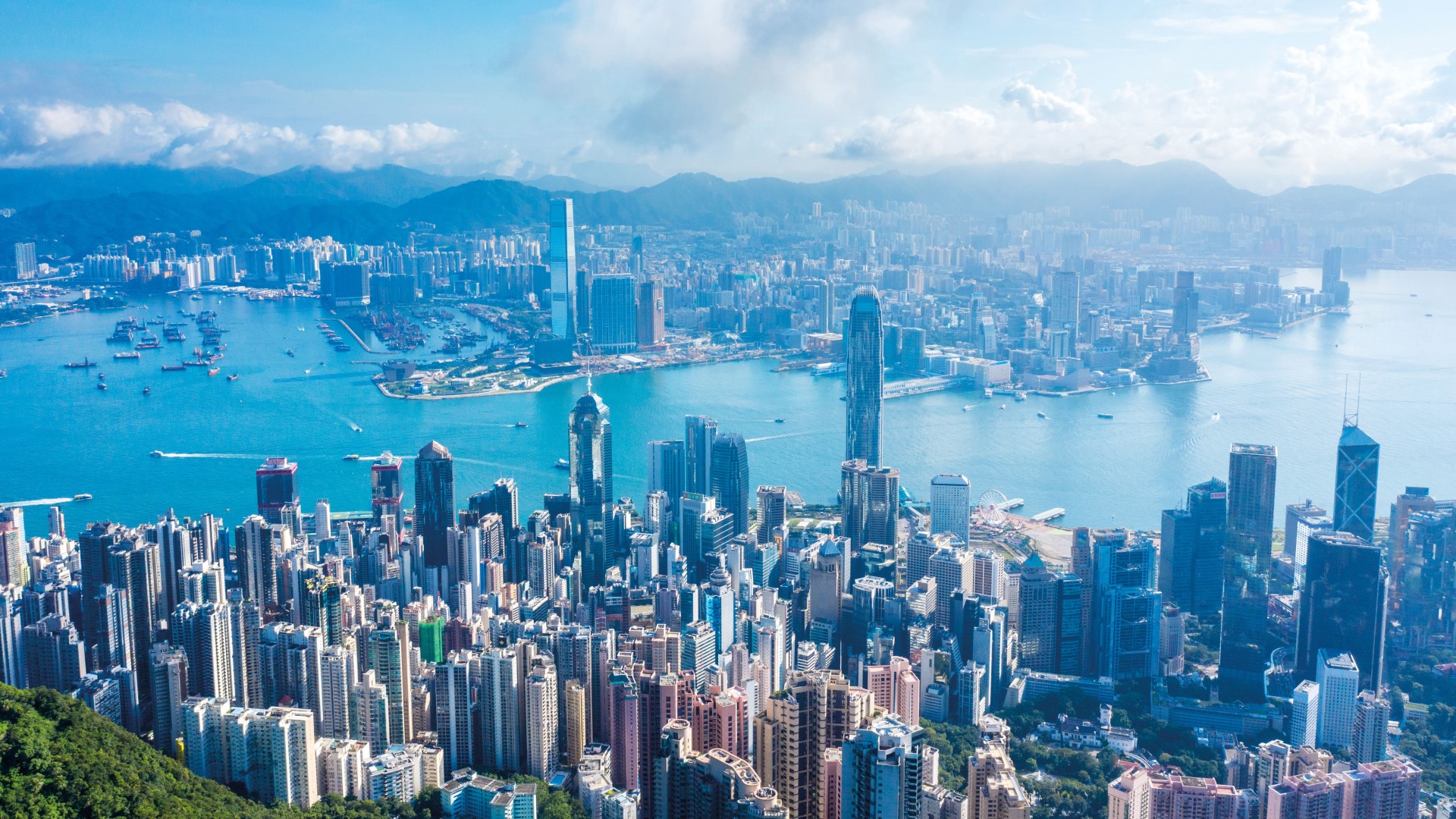}
    \end{minipage} \quad 
    \centering
    \begin{minipage}[f]{0.97\textwidth} 
    \vspace{2mm}
    \small
    {\color{prompt}\textbf{Prompt:}} Describe this image in detail. \\

    {\color{prompt}\textbf{Response:}} The image depicts a panoramic view of a bustling urban cityscape, likely a major metropolitan area. The foreground is dominated by numerous high-rise buildings, showcasing a dense concentration of skyscrapers. These buildings vary in height and architectural style, with some featuring modern glass facades and others displaying more traditional designs. The city appears to be a hub of economic activity, given the density and variety of the structures. \\

    In the midground, the city extends into a large body of water, likely a harbor or bay, with numerous boats and ships. This waterway is likely a significant transportation route, facilitating the movement of goods and people. The presence of numerous boats suggests a thriving maritime industry. \\
    
    The background features a vast expanse of water, likely a bay or a coastal area, with more high-rise buildings and urban development extending into the distance. The horizon is lined with a range of mountains or large hills, adding a natural backdrop to the otherwise man-made environment. The sky above is mostly clear with a few scattered clouds, indicating good weather conditions. \\
    
    The overall scene is one of economic prosperity and urban development, with the combination of modern skyscrapers and the expansive waterway suggesting a city that is both economically vibrant and strategically located for maritime trade. The presence of numerous boats and the extensive waterway indicate a well-developed infrastructure, likely supporting a significant population and economy. \\

    {\color{prompt}\textbf{Prompt:}} Where is this most likely to be? \\

    {\color{prompt}\textbf{Response:}} This image is most likely of Hong Kong. The presence of the distinctive twin-towered skyscrapers, the hilly terrain, the bay with numerous boats, and the urban skyline with numerous high-rise buildings are characteristic features of Hong Kong's skyline.

    \end{minipage}
\end{tcolorbox}

\centering
\begin{tcolorbox}[rounded corners, colback=white, colframe=black, left=5pt, right=5pt, top=5pt, bottom=5pt]
    \begin{minipage}[ft]{0.4\textwidth} 
        \includegraphics[width=\linewidth]{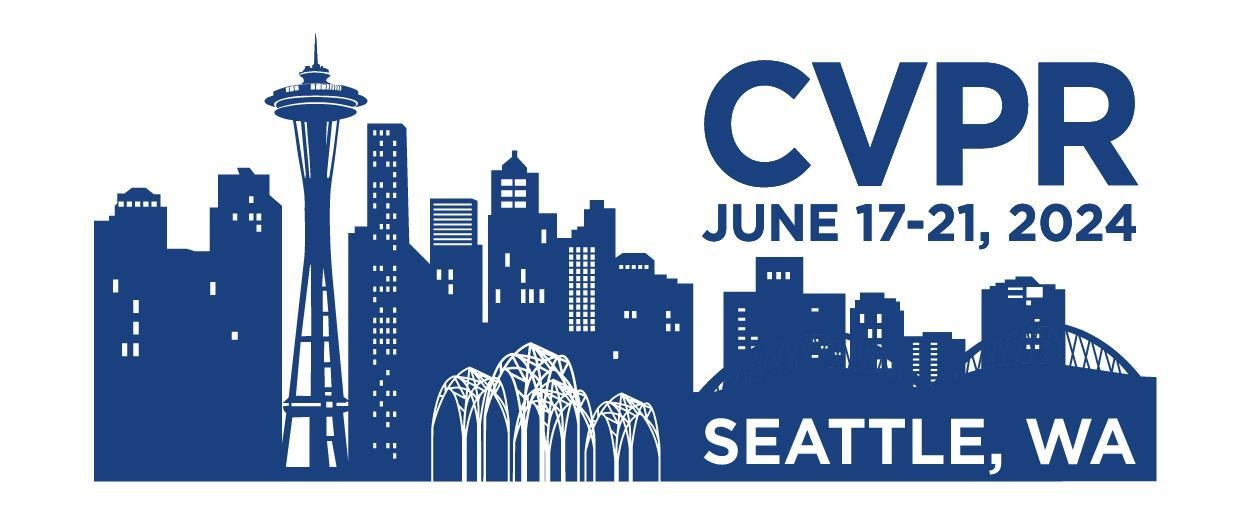}
    \end{minipage} \quad 
    \begin{minipage}[f]{0.56\textwidth} 
    \vspace{2mm}
    \small
    {\color{prompt}\textbf{Prompt:}} Tell me about this logo.  \\

    {\color{prompt}\textbf{Response:}} This is a logo for the CVPR conference, standing for the Conference on Computer Vision and Pattern Recognition, scheduled from June 17 to June 21, 2024, in Seattle, WA. The logo features a stylized blue cityscape of Seattle, prominently displaying notable landmarks such as the Space Needle. The text ``CVPR'' is displayed in large, bold letters to the right of the cityscape, with the dates and location information below it. The design is simple yet recognizable, effectively communicating the event's location and timing. \\
    \end{minipage}
\end{tcolorbox}

\end{figure*}

\begin{figure*}[ht]

    \centering
\begin{tcolorbox}[rounded corners, colback=white, colframe=black, left=5pt, right=5pt, top=5pt, bottom=5pt]
    \begin{minipage}[ft]{0.4\textwidth} 
        \includegraphics[width=\linewidth]{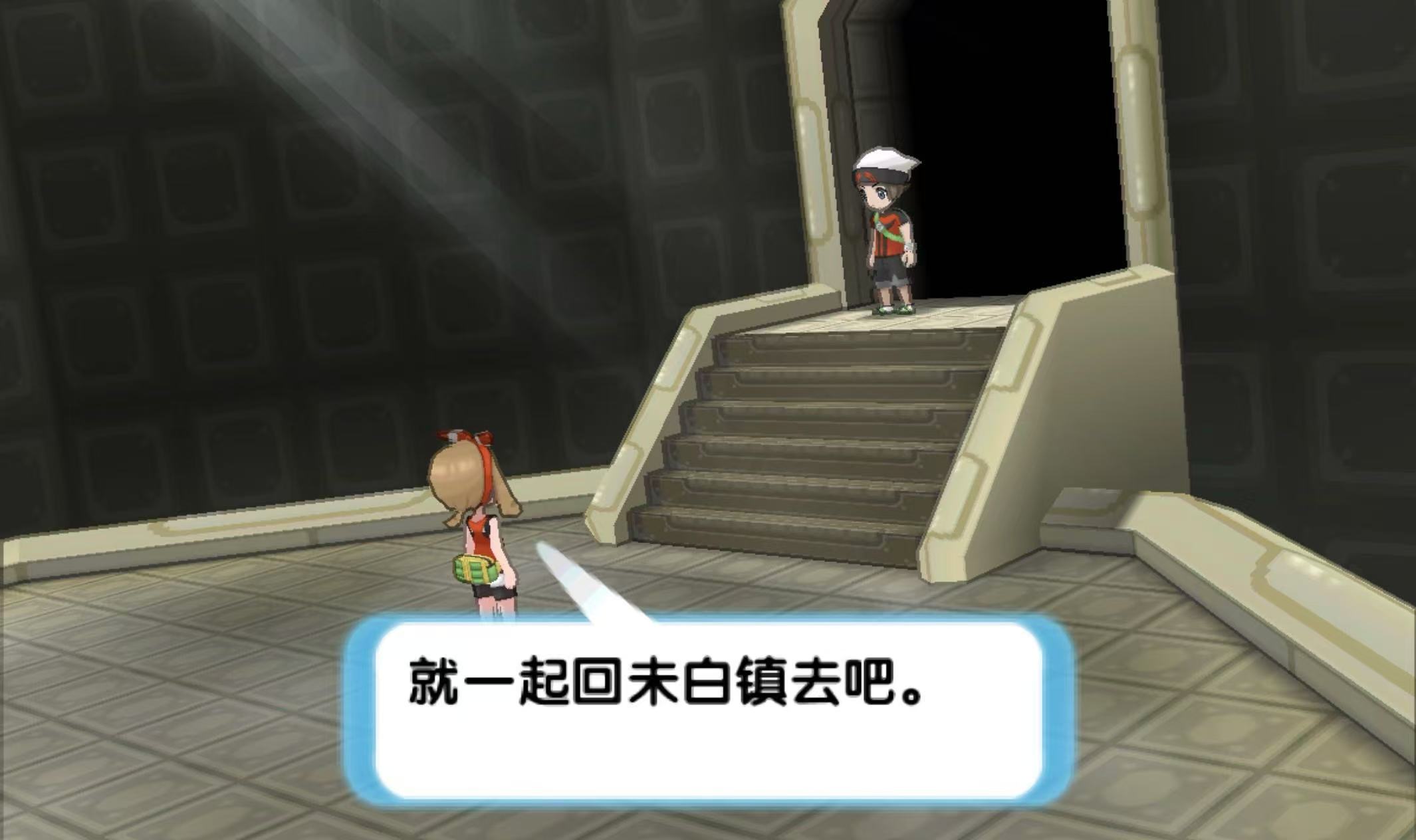}
    \end{minipage} \quad 
    \begin{minipage}[f]{0.56\textwidth} 
    \vspace{2mm}
    \small
    {\color{prompt}\textbf{Prompt:}} What is the content in this image?  \\

    {\color{prompt}\textbf{Response:}} The image appears to be a screenshot from a video game. The scene is set in a dimly lit, stone-walled environment that resembles a castle or fortress. The walls are made of large stone blocks, and there is a staircase leading up to a doorway on the right side of the image. The doorway is open, revealing a dark interior. \\
    
    In the foreground, there is a character with brown hair, wearing a red and white outfit. The character is facing away from the camera, looking up towards the stairs. \\
    
    In the background, another character is standing on the stairs, facing the camera. This character has a white cap, a red and white outfit, and is carrying a backpack. The character is looking down the stairs towards the open doorway.\\
    
    At the bottom of the image, there is a dialogue box with Chinese text. The text reads: ``\begin{CJK}{UTF8}{gbsn}就一起回未白镇去吧。\end{CJK}''. \\

    \end{minipage}
\end{tcolorbox}

\begin{flushleft}\textbf{OCR}\end{flushleft}
    \centering
\begin{tcolorbox}[rounded corners, colback=white, colframe=black, left=5pt, right=5pt, top=5pt, bottom=5pt]
    \begin{minipage}[ft]{0.3\textwidth} 
        \includegraphics[width=\linewidth]{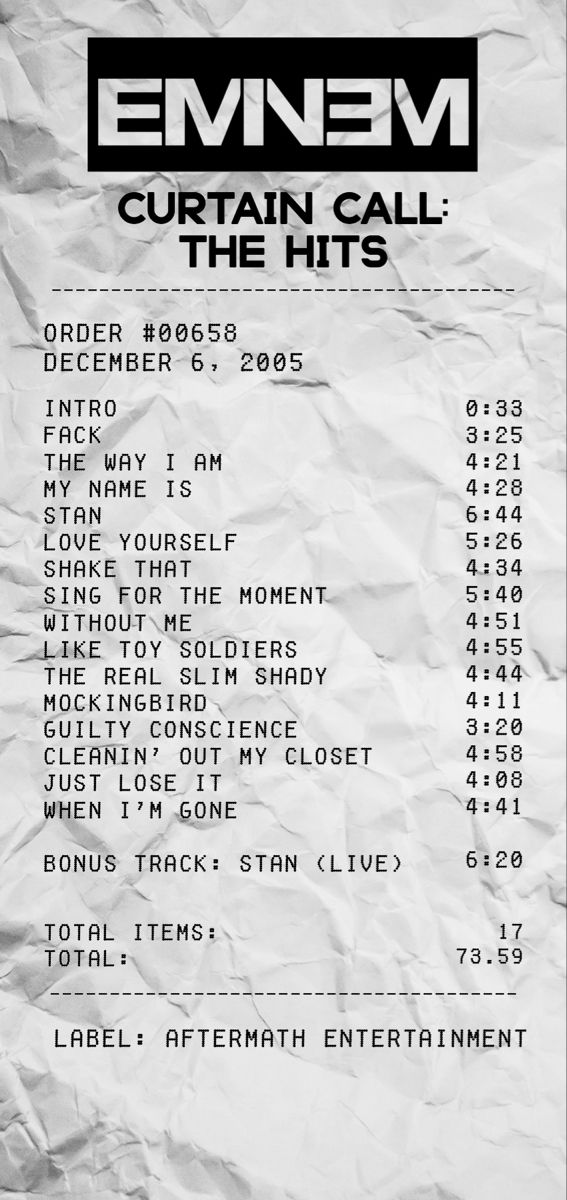}
    \end{minipage} \quad 
    \begin{minipage}[f]{0.56\textwidth} 
    \vspace{2mm}
    \small
    {\color{prompt}\textbf{Prompt:}} Describe this image from @albumreceipt.  \\

    {\color{prompt}\textbf{Response:}} This image appears to be a tracklist for a music release titled ``Curtain Call: The Hits'' by EMINEM, in a receipt-like style. The tracklist is as follows: \\

    1. Intro - 0:33  
    
    2. Fack - 3:25   
    
    3. The Way I Am - 4:21  
    
    4. My Name Is - 4:28  
    
    5. Stan - 6:44  
    
    6. Love Yourself - 5:26
    
    7. Shake That - 4:34
    
    8. Sing For The Moment - 5:40
    
    9. Without Me - 4:51
    
    10. Like Toy Soldiers - 4:55
    
    11. The Real Slim Shady - 4:44
    
    12. Mockingbird - 4:11
    
    13. Guilty Conscience - 3:20
    
    14. Cleanin' Out My Closet - 4:58
    
    15. Just Lose It - 4:08
    
    16. When I'm Gone - 4:41
    
    17. Stan (live, bonus track) - 6:20 \\
    
    The total tracklist includes 17 songs. The total length of the songs in this tracklist is 73:59. The release date is December 6, 2005. The label is ``Aftermath Entertainment''. \\

    \end{minipage}
\end{tcolorbox}
\end{figure*}

\begin{figure*}[ht]
\begin{flushleft}\textbf{Document Understanding}\end{flushleft}
    \centering
\begin{tcolorbox}[rounded corners, colback=white, colframe=black, left=5pt, right=5pt, top=5pt, bottom=5pt]
    \begin{minipage}[ft]{0.4\textwidth} 
        \includegraphics[width=\linewidth]{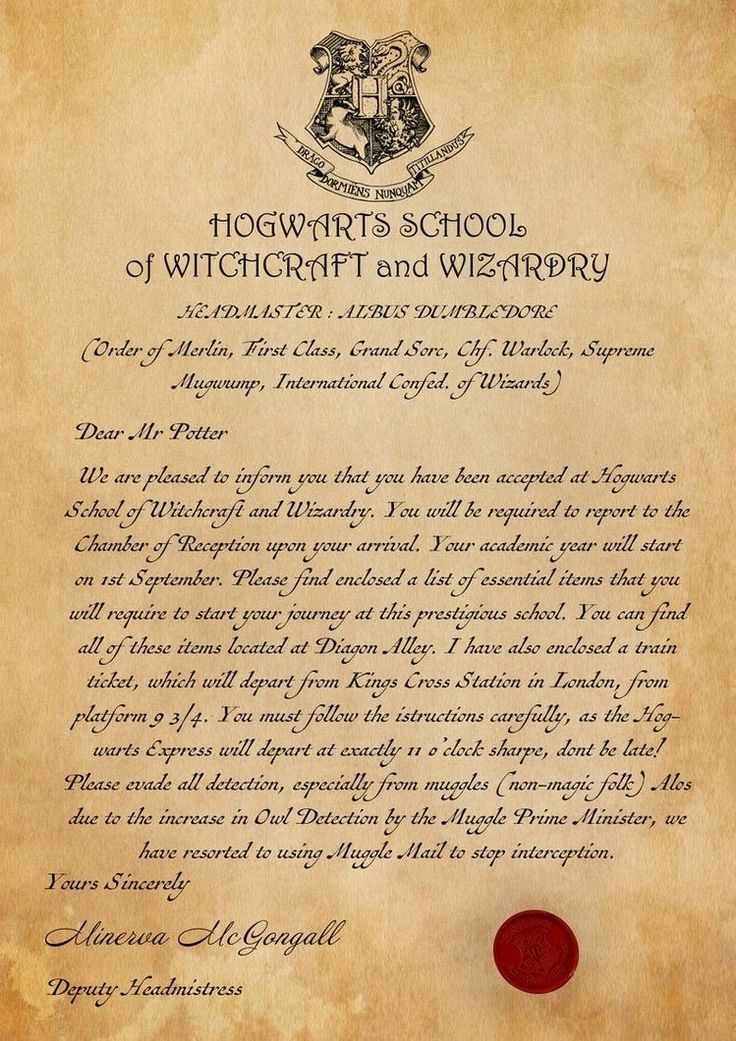}
    \end{minipage} \quad 
    \begin{minipage}[f]{0.56\textwidth} 
    \vspace{2mm}
    \small
    {\color{prompt}\textbf{Prompt:}} Which platform will the train depart from?  \\

    {\color{prompt}\textbf{Response:}} The train will depart from King's Cross Station in London, specifically Platform 9 3/4. \\

    {\color{prompt}\textbf{Prompt:}} Where should Mr. Potter buy the items needed for school?  \\

    {\color{prompt}\textbf{Response:}} Based on the image, Mr. Potter should buy the items needed for school from the Diagon Alley, which is mentioned as a requirement for the school trip. \\

        \end{minipage}
\end{tcolorbox}

\begin{flushleft}\textbf{Humor Understanding}\end{flushleft}
\centering
\begin{tcolorbox}[rounded corners, colback=white, colframe=black, left=5pt, right=5pt, top=5pt, bottom=5pt]
    \begin{minipage}[ft]{0.4\textwidth} 
        \includegraphics[width=\linewidth]{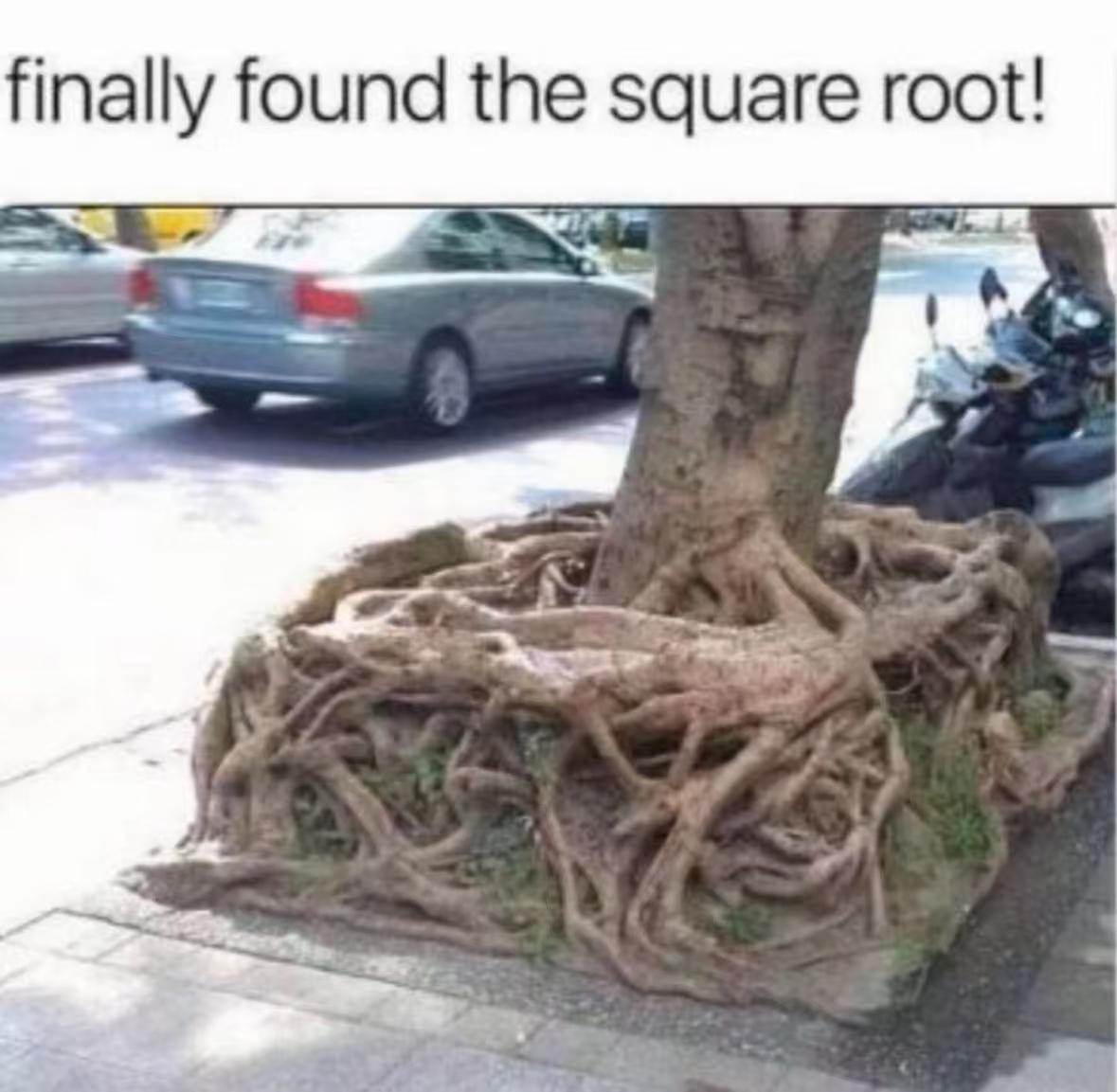}
    \end{minipage} \quad 
    \begin{minipage}[f]{0.56\textwidth} 
    \vspace{2mm}
    \small
    {\color{prompt}\textbf{Prompt:}} What is the punchline?  \\

    {\color{prompt}\textbf{Response:}} The punchline to the joke in the image is the unexpected and humorous twist on the common phrase ``finding the square root of a number.'' Instead of finding the square root, the image humorously shows someone finding a tree root in the square shape. The root structure is exaggerated to the point where it appears to be an infinite loop, adding a layer of absurdity and humor to the joke. \\

    \end{minipage}
\end{tcolorbox}

\end{figure*}

\begin{figure*}

\begin{flushleft}\textbf{Image to Latex}\end{flushleft}
    \centering
\begin{tcolorbox}[rounded corners, colback=white, colframe=black, left=5pt, right=5pt, top=5pt, bottom=5pt]

\begin{minipage}[ft]{0.9\textwidth} 
    \centering
    \includegraphics[width=0.6\linewidth]{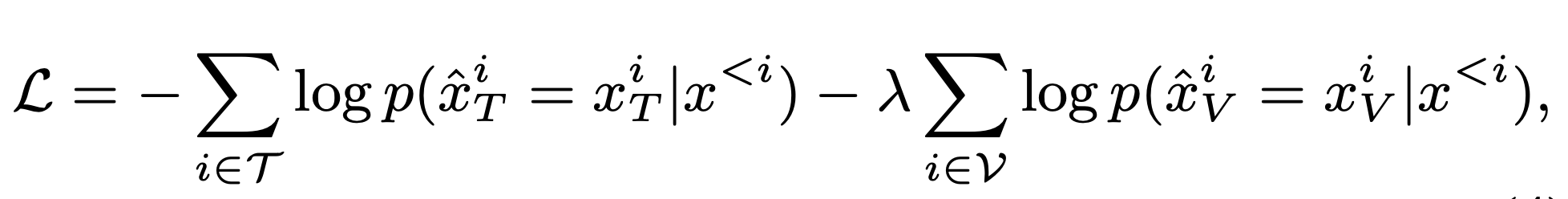}
    \vspace{1em}
\end{minipage}
    \centering
    \begin{minipage}[f]{0.97\textwidth} 
    \small
    
    {\color{prompt}\textbf{Prompt:}} Convert the formula in the figure to latex format.  \\
    
    {\color{prompt}\textbf{Response:}} Certainly, I'll convert the formula in the figure to LaTeX format for you.
    {
    \scriptsize
    \begin{verbatim}
    \begin{equation}
    \mathcal{L} = - \sum_{i \in \mathcal{T}} \log p(\hat{x}^i_T = x^i_T | x^{<i}) 
    - \lambda \sum_{i \in \mathcal{V}} \log p(\hat{x}^i_V = x^i_V | x^{<i}),
    \end{equation}
    \end{verbatim}
    }
    
\end{minipage}
\end{tcolorbox}

\begin{flushleft}\textbf{Math}\end{flushleft}
    \centering
\begin{tcolorbox}[rounded corners, colback=white, colframe=black, left=5pt, right=5pt, top=5pt, bottom=5pt]
    \begin{minipage}[ft]{0.3\textwidth} 
        \centering
        \includegraphics[width=0.8\linewidth]{figs/vis-understanding-math1.png}
    \end{minipage} \quad 
    \begin{minipage}[f]{0.66\textwidth} 
    \vspace{2mm}
    \small
    {\color{prompt}\textbf{Prompt:}} What is the perimeter of the rectangle?  \\

    {\color{prompt}\textbf{Response:}} To calculate the perimeter of the rectangle, we need to add up the lengths of all four sides. In this case, the rectangle has two sides that are each 6 feet long, and two sides that are each 3 feet long. \\

    The perimeter of a rectangle is calculated by adding the lengths of all four sides. So, the formula is: \\
    Perimeter = 2(length + width) \\
    
    For this rectangle, the length is 6 feet and the width is 3 feet. \\
    
    Plugging these values into the formula, we get: \\
    Perimeter = 2(6 + 3) = 2(9) = 18 \\
    
    Therefore, the perimeter of the rectangle is 18 feet. \\

    \end{minipage}
\end{tcolorbox}

\centering
\begin{tcolorbox}[rounded corners, colback=white, colframe=black, left=5pt, right=5pt, top=5pt, bottom=5pt]
    \begin{minipage}[ft]{0.3\textwidth} 
        \includegraphics[width=\linewidth]{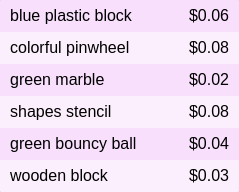}
    \end{minipage} \quad 
    \begin{minipage}[f]{0.66\textwidth} 
    \vspace{2mm}
    \small
    {\color{prompt}\textbf{Prompt:}} Lena has \$0.13. How much money will Lena have left if she buys a shapes stencil and a green bouncy ball? (Unit: \$)  \\

    {\color{prompt}\textbf{Response:}} To find out how much money Lena will have left after buying a shapes stencil and a green bouncy ball, we need to subtract the total cost of these items from the amount she has. \\

The total cost of the shapes stencil and the green bouncy ball is \$0.08 + \$0.04 = \$0.12. \\

To find out how much money Lena will have left, we subtract the total cost from the amount she has: \\

\$0.13 - \$0.12 = \$0.01 \\

Therefore, Lena will have \$0.01 left after buying a shapes stencil and a green bouncy ball. \\

    \end{minipage}
\end{tcolorbox}

\end{figure*}

\end{document}